\pdfoutput=1

\documentclass[11pt]{article}

\usepackage{acl}

\usepackage{times}
\usepackage{latexsym}

\usepackage[T1]{fontenc}

\usepackage[utf8]{inputenc}
\usepackage{tablefootnote}

\usepackage{microtype}
\usepackage{multirow}
\usepackage{multicol}
\usepackage{booktabs} 
\usepackage{changepage,threeparttable} 
\usepackage{makecell}
\usepackage{colortbl}
\usepackage{xspace}
\usepackage{enumitem}
\usepackage{dashrule}
\usepackage{arydshln}
\usepackage{hyperref}
\usepackage{url}
\usepackage{graphicx}
\usepackage{booktabs} 
\usepackage{changepage,threeparttable} 
\usepackage{natbib}
\usepackage[utf8]{inputenc} 
\usepackage[T1]{fontenc}    
\usepackage{url}            
\usepackage{booktabs}       
\usepackage{amsfonts}       
\usepackage{nicefrac}       
\usepackage{graphicx}
\usepackage{courier}
\usepackage{amsmath}
\usepackage{amssymb}
\usepackage{bbm}
\usepackage{adjustbox}
\usepackage{pifont} 
\usepackage{multirow}
\usepackage{multicol}
\usepackage{changepage,threeparttable} 
\usepackage{makecell}
\usepackage{colortbl}
\usepackage{xspace}
\usepackage{enumitem}
\usepackage{subfigure}
\usepackage{float}
\usepackage{wrapfig}
\usepackage{algorithm}
\usepackage{algpseudocode}
\usepackage{algorithmicx}

\usepackage{graphicx}
\usepackage[font=footnotesize,labelformat=simple]{subcaption}

\usepackage{booktabs}
\usepackage{changepage,threeparttable} 
\usepackage{natbib}
\usepackage[utf8]{inputenc} 
\usepackage[T1]{fontenc}    
\usepackage{url}            
\usepackage{amsfonts}       
\usepackage{nicefrac}       
\usepackage{courier}
\usepackage{amsmath}
\usepackage{amssymb}
\usepackage{bbm}
\usepackage{adjustbox}
\usepackage{pifont} 
\usepackage{multirow}
\usepackage{multicol}
\usepackage{array}
\usepackage{dashrule}
\usepackage{arydshln}
\usepackage{longtable}

\newcolumntype{L}[1]{>{\raggedright\arraybackslash}p{#1}}

\usepackage{makecell}
\usepackage{colortbl}
\usepackage{xspace}
\usepackage{enumitem}
\usepackage{float}
\usepackage{wrapfig}
\usepackage{algorithm}
\usepackage{geometry}
\usepackage{xcolor}
\usepackage{algorithmicx}
\usepackage{algpseudocode}
\usepackage{amsmath}

\title{SHARE: An SLM-based Hierarchical Action CorREction Assistant for Text-to-SQL}

\author{Ge Qu $^{1}$, Jinyang Li $^{1}$, Bowen Qin $^{2}$\thanks{\quad Corresponding authors are Bowen Qin and Reynold Cheng.} ,
Xiaolong Li $^{1}$, Nan Huo$^{1}$, \\ \textbf{Chenhao Ma} $^{3}$, \textbf{Reynold Cheng} $^{1}$\footnotemark[1]\\
$^{1}$The University of Hong Kong,
$^{2}$ BAAI\\
$^{3}$The Chinese University of Hong Kong, Shenzhen  \\
\texttt{quge@connect.hku.hk}, \texttt{bwqin@baai.ac.cn}, \texttt{ckcheng@cs.hku.hk} 
}

\begin{document}

\maketitle

\begingroup

\begin{abstract}

Current self-correction approaches in text-to-SQL face two critical limitations: 1) Conventional self-correction methods rely on recursive self-calls of LLMs, resulting in multiplicative computational overhead, and 2) LLMs struggle to implement effective error detection and correction for declarative SQL queries, as they fail to demonstrate the underlying reasoning path. In this work, we propose \textbf{SHARE}, an \textbf{S}LM-based \textbf{H}ierarchical \textbf{A}ction cor\textbf{RE}ction assistant that enables LLMs to perform more precise error localization and efficient correction. SHARE orchestrates three specialized Small Language Models (SLMs) in a sequential pipeline, where it first transforms declarative SQL queries into stepwise action trajectories that reveal underlying reasoning, followed by a two-phase granular refinement. We further propose a novel hierarchical self-evolution strategy for data-efficient training. Experimental results demonstrate that SHARE effectively enhances self-correction capabilities while proving robust across various LLMs. Furthermore, our comprehensive analysis shows that SHARE maintains strong performance even in low-resource training settings, which is particularly valuable for text-to-SQL applications with data privacy constraints. For reproducibility, we release our code at \url{https://github.com/quge2023/SHARE}.
\end{abstract}

\begin{figure}
    \centering
    \includegraphics[width=0.4\textwidth]{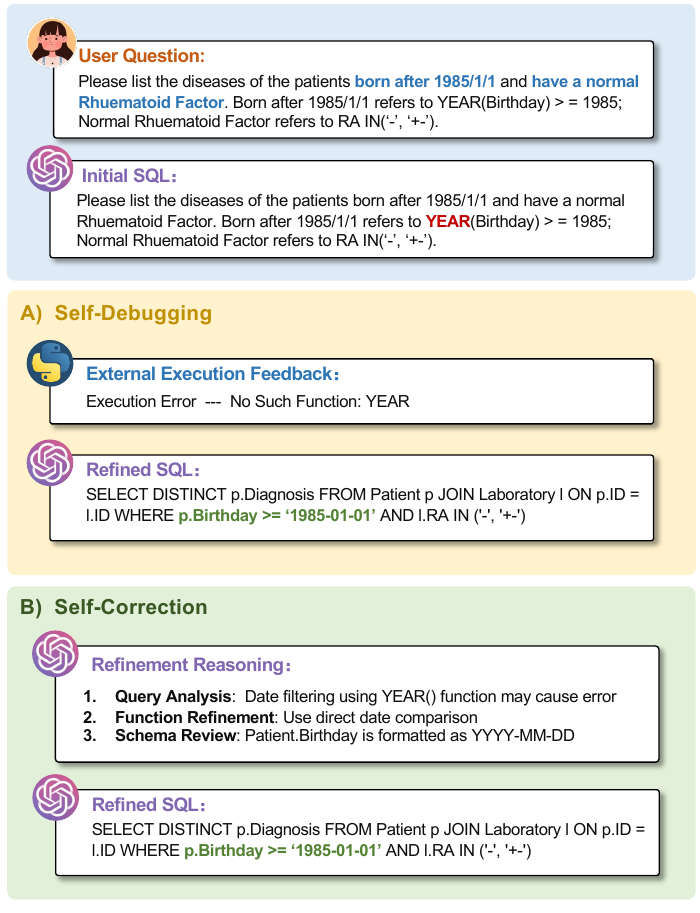}
    \caption{Illustrations of self-debugging and self-correction for text-to-SQL.}
    \label{fig:self-debug}
    \vspace{-0.2cm}
\end{figure}

\section{Introduction}

Text-to-SQL, aimed at converting natural language (NL) queries to executable SQL queries \citep{sun}, plays a crucial role in enabling non-technical users to analyze and interact with data in relational databases. In recent years, the application of Large Language Models (LLMs) has improved the performance of text-to-SQL to another level of intelligence \citep{evaluate_text_to_sql, chess, chase, table_reasoning}.
A critical component emerging from these architectural developments is the automatic error correction mechanism \citep{follow_feedback,codet,auto_correct_survey}, which systematically identifies and rectifies query errors to improve response accuracy.

Contemporary automatic error correction approaches can be categorized along two primary dimensions: self-debugging and self-correction. Self-debugging represents an execution-guided approach in which LLMs iteratively refine their SQL queries based on database execution feedback \citep{case_study_t2s, right_sql, decomposition_t2s}, as illustrated in Figure \ref{fig:self-debug} (A). While demonstrating promising results, this approach faces several fundamental challenges. First, execution feedback from mainstream SQL dialects such as SQLite tends to be concise but insufficiently specific, impeding accurate error localization. This ambiguity may subsequently induce hallucinations \citep{dialog_ha_1, ha_nlg} in the correction process. Furthermore, the fundamental requirement for direct database execution access presents significant operational constraints, particularly in contexts where data privacy and security considerations preclude such direct interaction with the database system \citep{data_privacy}.

These issues have motivated the focus of automatic correction towards self-correcting mechanisms \citep{epi-sql, magic}, where LLMs are prompted to revise their initial outputs through autonomous contextual re-analysis, without relying on external execution feedback. This decoupling from execution environments proves particularly valuable in text-to-SQL applications where database access is typically restricted by privacy constraints. However, achieving effective self-correction often requires multiple inference iterations through proprietary LLM APIs like GPT-4 or Claude-3.5-Sonnet, leading to prohibitive computational costs that scale exponentially. Furthermore, LLMs exhibit a tendency towards self-enhancement bias \citep{cannot_self_correct}, leading them to overestimate the quality of their initial outputs and struggle to effectively identify errors within their self-generated declarative SQL queries. This inherent limitation undermines the overall effectiveness of the correction process.

\begin{figure*}[t]
    \centering
    \includegraphics[width=1.0\textwidth]{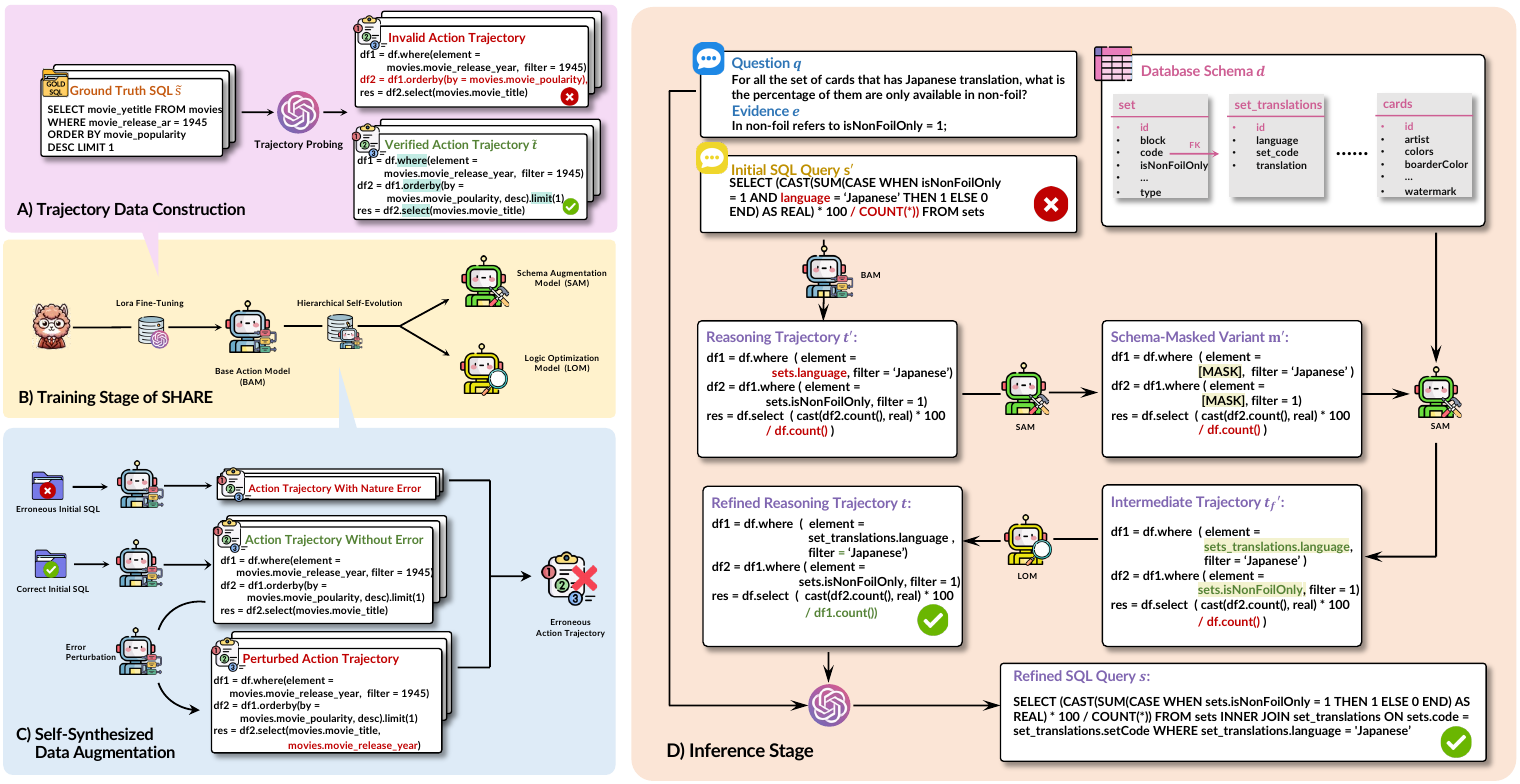}
    \caption{An illustration of SHARE. Figure (A)-(C) illustrate the training architecture of three specialized SLMs in SHARE: the Base Action Model (BAM), Schema Augmentation Model (SAM), and Logic Optimization Model (LOM). Figure D presents SHARE's orchestration of these three models in a sequential pipeline for inference.}
    \label{fig:main_figure}
\end{figure*}

In this work, we propose an assistant-based framework where generator LLMs create initial outputs and implement self-correction guided by assistants. Our primary contribution, \textbf{SHARE} (\textbf{S}LM-based \textbf{H}ierarchical \textbf{A}ction Cor\textbf{RE}ction Assistant), orchestrates three specialized Small Language Models (SLMs), each under 8B parameters, in a sequential pipeline. Specifically, the \textbf{B}ase \textbf{A}ction \textbf{M}odel (BAM) transforms raw SQL queries into action trajectories that capture reasoning paths; the \textbf{S}chema \textbf{A}ugmentation \textbf{M}odel (SAM) and the \textbf{L}ogic \textbf{O}ptimization \textbf{M}odel (LOM) further perform orchestrated inference to rectify schema-related and logical errors, respectively, within action trajectories. SHARE improves error detection precision and correction efficacy while reducing computational overhead compared to conventional LLM approaches. Additionally, we also incorporate a novel \textbf{hierarchical self-evolution strategy} that enhances data efficiency during training.

Experimental results across 4 diverse text-to-SQL benchmarks demonstrate the effectiveness, efficiency, and generalizability of SHARE. By enabling SLMs to collaboratively guide LLMs in SQL correction, SHARE achieves substantial execution accuracy improvements over the GPT-4o baseline, with relative gains of 14.80\% on BIRD and 11.41\% on SPIDER within a single round of correction, while significantly reducing computational costs (Section \ref{sec:cost}). Beyond precision, our framework exhibits strong robustness in varying query complexities, low-resource settings, and various generator models, including both closed-source and open-source LLMs (Section \ref{sec:overall}). Notably, SHARE’s learned debugging logic generalizes effectively to previously unseen SQL dialects without dialect-specific supervision (Section \ref{sec:dialect}). These findings position our method as a scalable and cost-efficient paradigm for improving the reasoning reliability of LLMs in SQL auto-correction for real-world applications.

\section{Preliminaries}
\subsection{Task Definition}
\paragraph{Text-to-SQL.}
Given a natural language question $q_i \in \mathcal{Q}$, where $\mathcal{Q} = \left\{q_i\right\}_{i=1}^n$ with its corresponding database input \footnote{Database input refers to the schema of a specific database and its corresponding sampled value.} $d_i \in \mathcal{D}$, where $\mathcal{D} = \left\{d_i\right\}_{i=1}^n$, the goal of text-to-SQL is to guide the generator model $\mathcal{G}$ to generate the SQL query $s_i$ by:
\begin{equation}
s_i = f_{\mathcal{G}}(d_i,q_i),
\label{eq1}
\end{equation}
where $f_{\mathcal{G}}\left(\cdot \right)$ refers to the mapping function applied by the generator model $\mathcal{G}$.

\paragraph{Assistant-based Self-Correction.}
Given the $d_i$, $q_i$ and the corresponding initial SQL queries $s'_i$ generated by the generator model $\mathcal{G}$, the goal of assistant-based correction aims to utilize feedback $y_i$ generated by the assistant $\mathcal{T}$, which could be a LLM or an agent mechanism, to guide the generator model $\mathcal{G}$ to refine its initial output effectively by:
\begin{equation}
s_i = f_{\mathcal{G}}(d_i, q_i, s'_i, y_i),
\label{eq2}
\end{equation}
where $y_i = f_{\mathcal{T}}(d_i, q_i, s'_i)$, and $f_{\mathcal{T}}\left(\cdot \right)$ is the mapping function applied by the assistant. The generator model $\mathcal{G}$ and assistant $\mathcal{T}$ engage in multiple iterative refinement cycles until a correct SQL query is produced or the predefined maximum number of turns is reached.

\subsection{Action Model}
An action model \citep{action_model} $\mathcal{A}$ is specifically architected to comprehend, plan, and generate stepwise action trajectories $t_i$, given the contextual input $x_i$ by:
\begin{equation}
t_i = \{a_1, a_2, ..., a_n\} = f_{\mathcal{A}}(x_i ; \mathbb{A}),
\label{eq3}
\end{equation}
where $f_{\mathcal{A}}\left(\cdot ; \mathbb{A} \right)$ refers to the mapping function applied by the action model $\mathcal{A}$ with an action space $\mathbb{A}$ inherently mastered by $\mathcal{A}$, and $a_i \in \mathbb{A}$ denotes as the selected action. Unlike natural language, actions manipulated by action models are typically in the format of API or function calls. For instance, a declarative SQL query \texttt{"SELECT .. FROM .. WHERE .."} can be decomposed by an action model into a sequence of functional operations as \texttt{[where(column = param\_1, value = param\_2), select(table = param\_1, column = param\_2)]}.

\begin{table*}[t]
\centering
\footnotesize
\begin{tabular}{L{2cm}p{4.2cm}p{8cm}}
\toprule
{\bf Type} & {\bf Definition} &{\bf Example}\\
 \midrule 
ADD & 
Inserts an additional action into the original action trajectory. 
\newline
\texttt{$\left[a_1, a_2, ...,a_n\right] \to$}
\newline
\texttt{$\left[ a_1, a_2,\boldsymbol{\textcolor{purple}{a_{new}}},...,a_n \right]$} &
\textbf{Before:} 
\newline
df1 = df.orderby(element = movie.likes, desc).limit(1)
\newline
res = df1.select(element = movie.director)
\newline
\textbf{After:} 
\newline
\textbf{\textcolor{purple}{df1 = df.groupby(element = movies.id)}}
\newline
df2 = df1.orderby(element = movie.likes, desc).limit(1)
\newline
res = df2.select(element = movie.director)\\
\midrule

DELETE & 
Removes an existing action from the trajectory.
\newline
\texttt{$\left[a_1, \boldsymbol{\textcolor{teal}{a_2}}, a_3,...,a_n\right] \to$}
\newline
\texttt{$\left[ a_1, a_3,...,a_n \right]$} &
\textbf{Before:} 
\newline
df1 = df.where(element = users.country\_code,filter = 20)
\newline
\textbf{\textcolor{teal}{df2 = df1.where(element = users.gender, filter = 'male')}}
\newline
res = df2.select(users.user\_id)
\newline
\textbf{After:} 
\newline
df1 = df.where(element = users.country\_code,filter = 20)
\newline
res = df1.select(users.user\_id)\\
\midrule

SUBSTITUTE & 
Replaces an existing action with a different action type or modifies the parameters of the existing action. 
\newline
\texttt{$\left[a_1, \boldsymbol{\textcolor{purple}{a_2}}, a_3,...,a_n\right] \to$}
\newline
\texttt{$\left[a_1, \boldsymbol{\textcolor{teal}{a'_2}}, a_3,...,a_n\right] \to$} &
\textbf{Before:} 
\newline
df1 = df.where(element = reviews.Date, filter = '2018-09-11')
\newline
res = df1.\textbf{\textcolor{teal}{select(district.district\_id, district.city)}}
\newline
\textbf{After:} 
\newline
df1 = df.where(element = reviews.Date, filter = '2018-09-11')
\newline
res = df1.\textbf{\textcolor{purple}{select(district.city, district.district\_id )}}\\
 \bottomrule
\end{tabular}
\caption{Three error perturbation types utilized by Base Action Model (BAM) to implement data augmentation.}
\label{tab:perturbation}
\end{table*}

\section{Methodology}
\subsection{Training Pipeline}
The general training data used to develop SLMs is often generated through knowledge distillation pipelines, thereby reducing the need for costly human expert annotations \citep{kd_survey, enhanced_kd}. In these pipelines, LLMs serve as \textbf{teacher} models, providing annotations and generating new training instances to guide the learning process of SLMs as \textbf{student} models. We employ this approach to train a Base Action Model (BAM), which transforms initially produced declarative SQL queries into structured sequences of actions within a predefined action space, capturing the underlying reasoning steps. Building on this foundation, two action models specialized for two crucial aspects of text-to-SQL are further presented: the Schema-Augmentation Model (SAM) and the Logic Optimization Model (LOM). SAM concentrates on improving schema linking \citep{uni_sar}, while LOM focuses on logical synthesis \citep{logical_synthesis}. All our models were trained via Lora fine-tuning \citep{pure_lora}. 

To train SAM and LOM efficiently, we propose a novel continual learning technique: \textbf{hierarchical self-evolve strategy}. Instead of repeatedly querying a teacher LLM (e.g., GPT-4o) for each new training instance, this strategy leverages BAM to synthesize and augment task-specific training data, targeting distinct aspects of text-to-SQL translation. This approach reduces annotation costs while maintaining strong performance, as demonstrated in Section \ref{sec:cost}. Formally, we begin with a seed dataset $\mathcal{C} = \{(d_i, q_i, \tilde{s}_i, s’_i)\}_{i=1}^n$, where each tuple $(d_i, q_i, \tilde{s}_i)$ consists of a database input, a user question, and the associated ground-truth SQL from the BIRD and SPIDER training corpora. A detailed distribution of the training data is provided in Appendix \ref{app:training_data_dist}. The initial SQL $s_i'$ is generated by GPT-4o using the baseline prompt implemented in BIRD \citep{bird}. 

\subsection{Base Action Model (BAM)}
\paragraph{Training Target.}
\label{sec:bam}
 BAM aims to generate the corresponding action trajectory $t$ given an initial SQL query $s'$. Following prior study \citep{ta-sql}, we design actions in target trajectories as pandas-like APIs (see Refined Reasoning Trajectory in Figure \ref{fig:main_figure} (D)) to present the reasoning process of text-to-SQL transformations. The complete enumeration of actions used in the construction of action trajectories within the SHARE framework is detailed in the Appendix \ref{app:actions}.

\paragraph{Data Construction.}
Given that BAM is the most important model and such reasoning derivations are complex and highly demanding, we employ GPT-4o as a strong teacher LLM to construct the training data for BAM. As shown in Figure \ref{fig:main_figure} (A), GPT-4o is guided to convert each ground-truth (gt) SQL $\tilde{s}$ to a verified action trajectory $\tilde{t}$ by few-shot prompting. The prompt we use is detailed in Appendix \ref{app:bam_data_prompt}. To ensure high-quality training data with fewer hallucinations during conversion, we only contain $(\tilde{s}, \tilde{t})$ pairs in which $\tilde{t}$ can be successfully reverted back to $\tilde{s}$ \citep{reverse}. We ultimately collected 13k query-trajectory pairs as the training data for BAM in this phase. 

\subsection{Schema Augmentation Model (SAM)}
\paragraph{Training Target.}
Schema linking is a critical step in identifying the relevant database tables and columns needed to answer user queries \citep{sl_role}. However, the complexity and heterogeneity of database schemas, spanning from pre-trained language models (PLMs) \citep{plm, graphix} to large language models (LLMs) \citep{llm_survey}, often compromise accurate schema linking, thereby introducing substantial burdens on downstream SQL generation. To address this issue, we propose the Schema Augmentation Model (SAM), designed to specifically target and correct schema-linking errors within the input action trajectory. By focusing on schema-related components, SAM aims to isolate and rectify errors before they propagate, ultimately leading to more reliable SQL generation.

\paragraph{Data Construction.}
We begin with a corpus of 13k action trajectories from BAM, each represented as $(d, q, s', \tilde{t})$. For each verified trajectory $\tilde{t}$, we generate a schema-masked variant $\tilde{m}$ by inserting mask symbols \texttt{[MASK]} to isolate schema-specific elements. For each initial SQL $s'$, we extract its referenced tables and columns as an initial schema list $l'$. These paired forms, $(\tilde{t}, \tilde{m})$ and $(d, q, l', \tilde{m}, \tilde{t})$, form the backbone of SAM’s two-phase training paradigm. In the first phase, SAM is trained with $(\tilde{t}, \tilde{m})$ to identify and mask schema-related elements accurately. In the second phase, SAM leverages $(d, q, l’, \tilde{m}, \tilde{t})$ to refine the previously masked segments, seamlessly reintegrating the corrected schema links into the schema-masked variant. By employing BAM in a few-shot setting to prepare and orchestrate these training steps, we ensure that SAM efficiently acquires the specialized capabilities needed for robust schema augmentation.

\subsection{Logic Optimization Model (LOM)}
\label{sec:lom}
\paragraph{Training Target.}
Given a more precise set of schema-linked database tables or columns, the reasoning logic expressed as an action trajectory should align with both natural language descriptions and valid SQL semantics. Formally, the model takes $d$ and $q$ as input and outputs a refined action trajectory $t$. This trajectory $t$ captures the correct chain of reasoning necessary to accurately resolve the question $q$.

\paragraph{Data Construction.}
As shown in Figure \ref{fig:main_figure} (C), the erroneous action trajectories in the training data for LOM come from two resources. First, we collect the corresponding action trajectory of erroneous initial SQLs. However, the scale of this resource is limited. Therefore, we propose an \textbf{action-based perturbation strategy} for data augmentation. We apply three types of perturbation, as illustrated in Table \ref{tab:perturbation}, on error-free action trajectories derived from correct initial SQL to reproduce various logic errors in text-to-SQL conversions. Finally, 15k erroneous action trajectories along with their corresponding verified action trajectories are collected for LOM training. The detailed pseudocode of this process is provided in Appendix \ref{app:ps_code}.

\subsection{Orchestration Inference}
During inference, as shown in Figure \ref{fig:main_figure} (D), SHARE operates in a sequential orchestration that integrates three LoRa fine-tuned models, enabling iterative refinement of action trajectories. Upon receiving an initial SQL query $s’$, SHARE first invokes the BAM to generate a corresponding action trajectory $t’$. The SAM then refines $t’$ by applying schema-based adjustments derived from the given database input $d$, producing the intermediate trajectory $t_f’$. Next, $t_f’$, along with the user query and the refined database content, is forwarded to the LOM for logic-based corrections. The refined trajectory $t$ generated by LOM serves as SHARE’s final output and is employed as feedback to guide the underlying language model in self-correcting $s’$ within a zero-shot setting. Ultimately, with these refined action trajectories serving as self-correction signals, the language model can regenerate more accurate and contextually appropriate SQL queries.

\section{Experiments}
\subsection{Experiment Settings}

\paragraph{Datasets and Metrics.} 
The experiments are conducted on four challenging benchmarks for cross-domain text-to-SQLs.  1) \textbf {\textsc{Bird}} \citep{bird} is the most challenging large-scale cross-domain text-to-SQL benchmark, which introduces external knowledge as an additional resource in complex scenarios. In this paper, we use its development set for evaluation, which contains 1,534 pairs of text-to-SQL data and 11 complex databases. 2) \textbf{\textsc{Spider}}\citep{spider} is a more standard cross-domain text-to-SQL benchmark. It contains 1,034 examples, covering 20 databases across multiple domains, in the development set. 3) \textbf{\textsc{DK}} \citep{dk}, extended from the SPIDER benchmark, requires text-to-SQL parsers equipped with the ability of domain-knowledge reasoning. 4) \textbf{\textsc{Realistic}} removes and switches the obvious mentions of schema items in questions, making it closer to the real scenarios. In this paper, we use widely adopted \textbf{Execution Accuracy} (EX) to measure the performance of our framework.

\paragraph{Compared Methods.} 
We explore two open-source SLMs, namely Llama-3.1-8B and Phi-3-Mini-3.8B, as backbone models to construct our SHARE. For all baseline and advanced self-correction methods for comparison in Table \ref{tab:mainresult}, we employ GPT-4o as the generator model and report the results after a single refinement iteration. Details of these methods are shown in Appendix \ref{app:baselines}.

\paragraph{Implementation Details.}
We fine-tune all our models using the LLaMa-Factory library \citep{llamafactory} with LoRA \citep{pure_lora}. All our experiments are conducted on 4\(\times\)A100 GPU with 80GB memory. We detail the hyperparameters for training and inference in Appendix \ref{app:hyper_para}, and claim the reproducibility of this work in Appendix \ref{app:reproduce}.

\subsection{Overall Performance}
\label{sec:overall}
\begin{table*}[t]
    \centering
    \resizebox{1.0\textwidth}{!}{
        \begin{tabular}{lcccc cccccc}
            \toprule
            \multirow{2}{*}{\textbf{Method}} & \multicolumn{4}{c}{\textbf{\textsc{Bird}}} & \multicolumn{5}{c}{\textbf{\textsc{Spider}}} \\
            \cmidrule(lr){2-5} \cmidrule(lr){6-10}
            & \textbf{Simple} & \textbf{Moderate} & \textbf{Challenging} & \textbf{Total} & \textbf{Easy} & \textbf{Medium} & \textbf{Hard} & \textbf{Extra Hard} & \textbf{Total} \\ 
            \midrule
            GPT-4o & 63.35 & 44.18 & 45.52 & 55.87 & 89.10 & 83.00 & 68.40 & 52.40 & 77.10 \\
            \hline
            \rowcolor{black!10!}\multicolumn{10}{c}{\textbf{\textit{Self Correction w/o Feedback}}} \\
            Self-Correction & 62.70 & 43.75 & 44.83 & 55.28 & 88.30 & 82.70 & 66.10 & 49.40 & 75.90 \\
            
            Self-Consistency & 65.75 & 49.04 & 45.21 & 58.75 & 92.30 & 87.90 & 75.10 & 58.60 & 81.80 \\
            Multiple-Prompt & 66.38 & 48.06 & 44.83 & 58.80 & 91.10 & 87.20 & 74.70 & 59.00 & 81.50 \\
            \hline
            \rowcolor{black!10!}\multicolumn{10}{c}{\textbf{\textit{Self Correction w/ Feedback}}} \\
            Self-Debugging $^\clubsuit$ & 65.41 & 47.84 & 46.21 & 58.28 & 91.10 & 85.90 & 74.70 & 60.80 & 81.20 \\
            DIN-Correction & 65.62 & 46.98 & 44.83 & 58.02 & 91.90 & 85.20 & 70.10 & 55.40 & 79.50\\
            MAC-Refiner $^\clubsuit$& 66.27 & 47.41 & \underline{46.90} & 58.74 & \underline{93.10} & 85.00 & 73.00 & 56.60 & 80.40\\
            MAGIC & 66.75 & 49.46 & 45.79 & 59.53 & - & - & - & - & \underline{85.66}\\
            \hline
            \textbf{+SHARE-3.8B} & \underline{68.00} & \underline{51.29} & 46.21 & \underline{60.89} & 92.70 & \underline{88.30} & \underline{78.70} & \underline{65.10} & 84.00\\
            \textbf{+SHARE-8B} & \textbf{70.81} & \textbf{56.25} & \textbf{46.90} & \textbf{64.14} & \textbf{94.00} & \textbf{90.10} & \textbf{78.20} & \textbf{70.50} & \textbf{85.90}\\
            \bottomrule
        
        \end{tabular}
    }
    \caption{Self-correction performance of GPT-4o in Execution Accuracy (EX) (\%) on \textsc{BIRD} and \textsc{SPIDER}. $^\clubsuit$ means the model uses external execution results as feedback. \textbf{Bold} indicates best results, while \underline{underlines} denote second-best results.}
    \label{tab:mainresult}
\end{table*}

\paragraph{Overall Results.}

Table \ref{tab:mainresult} presents the performance of GPT-4o on the BIRD and SPIDER benchmarks, comparing approaches with baseline and advanced self-correction strategies. Three key observations emerge:
1) When GPT-4o attempts intrinsic Self-Correction, performance actually decreases (55.87\% $\to$ 55.28\% on BIRD). This occurs because the model lacks reliable mechanisms to assess the correctness of its prior reasoning steps, sometimes converting originally correct solutions into incorrect ones \citep{cannot_self_correct}. Moreover, improving prompts for self-correction strategies can inadvertently introduce biases, leading to suboptimal revisions. \citep{critic}. Although advanced methods with carefully crafted designs, such as Multiple-Prompt and Magic, mitigate these biases, they rely heavily on extensive human-engineered prompts and entail significant computational overhead.
2) In contrast, SHARE enables GPT-4o to perform effective self-correction through a single interaction, resulting in a relative improvement of 14.80\% in EX on BIRD and 11.41\% on SPIDER with SHARE-8B. That is because SHARE introduces a novel mechanism for inferring and analyzing initial hidden reasoning paths, allowing the LLM to identify and rectify errors more precisely. Notably, this process is conducted automatically by SLMs, thereby reducing both the human effort and the computational costs associated with high-level LLMs.
3) Significant performance improvements using SHARE-3.8B and SHARE-8B, which are based on two widely used SLMs (Phi-3-mini and Meta-Llama-8B), demonstrate the generalization and robustness of our training pipeline.
\begin{table}[t]
\centering
\resizebox{0.9\hsize}{!}{
\begin{tabular}{lccccc}
\toprule
\textbf{\textsc{Method}}    & {\textbf{\textsc{easy}}}      &  {\textbf{\textsc{med.}}}   
&  {\textbf{\textsc{hard}}} &  {\textbf{\textsc{extra}}}
&  {\textbf{\textsc{all}}}\\
\hline
 \rowcolor{black!10!}\multicolumn{6}{c}{\textbf{\textit{DK}}} \\

GPT-4o & 75.50&73.60&47.30&41.90&64.10 \\
+SHARE-3.8B &84.90&73.80&48.90&56.20&69.20 \\
+SHARE-8B& 85.50&81.30&56.80&63.80&75.30\\
\hline
 \rowcolor{black!10!}\multicolumn{6}{c}{\textbf{\textit{REALISTIC}}} \\

GPT-4o &81.70&82.30&69.70&49.50&73.40 \\
+SHARE-3.8B &88.10&86.70&68.70&57.70&78.00 \\
+SHARE-8B &87.20&88.20&74.70&68.00& 81.50\\
                                      
\bottomrule
\end{tabular}}
\caption{Execution Accuracy ({EX}) of GPT-4o + SHARE across queries of varying levels of difficulty on \textsc{DK} and \textsc{Realistic}.}
\label{tab:spider_res}
\end{table}

\begin{table}[t]  
    \centering
    \resizebox{0.9\hsize}{!}{
    \begin{tabular}{lcccc}  
    \toprule
    \textbf{\textsc{Model}}& \textbf{\textsc{sim.  }} & \textbf{\textsc{mod.  }} & \textbf{\textsc{chall.}} & \textbf{\textsc{total}} \\ 
    \hline
    \rowcolor{black!10!}\multicolumn{5}{c}{\textbf{\textit{Closed-Source LLM}}} \\
    Claude-3.5-S &57.08&39.87&31.03&49.41\\
    +SHARE-8B &68.86&57.11&50.34&63.56\\
    \midrule
    GPT-4o-mini &55.03&41.59&35.17&49.09\\
    +SHARE-8B &67.46&50.86&40.00&59.64\\
    \midrule
    GPT-3.5-turbo &52.51&35.99&29.66&45.35\\
    +SHARE-8B &57.51&38.36&31.03&49.22\\
    \hline
    \rowcolor{black!10!}\multicolumn{5}{c}{\textbf{\textit{Open-Source weaker LM}}} \\
    Llama-3.1-70B &60.54&44.61&41.38&53.91\\
    +SHARE-8B &68.86&53.88&43.45&61.93\\
    \midrule
    Qwen-Coder-32B &65.72&46.71&45.23&58.03\\
    +SHARE-8B &67.78&54.31&46.90&61.73\\
    \midrule
    Llama-3.1-8B &41.62&27.59&20.00&35.33\\
    +SHARE-8B &47.68&35.13&28.97&42.11\\
    \midrule
    DS-Coder-6.7B &41.30&26.94&23.45&34.57\\
    +SHARE-8B &56.25&45.04&36.55&51.24\\
    \bottomrule
    \end{tabular}}
    \caption{Self-correction performance of various generator models assisted by SHARE on \textsc{BIRD}. For brevity, we refer to some models using shorthand names and provide their corresponding official model aliases in Appendix \ref{app:full_name}. \textsc{sim.}, \textsc{mod.}, \textsc{chall.} represent the levels of query difficulty and are the abbreviations of simple, moderate, and challenging, respectively.}
    \label{tab:model_ad}
\end{table}

\paragraph{Results on Robust Testing.}

Table \ref{tab:spider_res} presents evaluation results in EX on the variant datasets of SPIDER for robustness without any additional training. SHARE-8B effectively enhances the performance of the GPT-4o baseline by 11.20\% and 8.10\% on DK and REALISTIC, respectively, with improvements across all difficulty levels. SHARE-3.8B also facilitates significant performance gains, achieving a relative increase of 7.96\% on DK and 7.18\% on REALISTIC.

\paragraph{Results on Various Generator Models.} 

Although the teacher LLMs we employ for data generation during training are primarily based on GPT-4o, the performance gains observed with our trained SHARE-8B assistant extend well beyond this single source, as evidenced in Table~\ref{tab:model_ad}. Notably, our approach significantly improves performance for both proprietary closed-source models, such as Claude-3.5-Sonnet ($\uparrow$ 28.64\% relatively), and for open-source alternatives, such as Llama-3.1-70B ( $\uparrow$ 14.88\% relatively). This indicates that our method is not limited to the model bias of error patterns in text-to-SQL, but also general knowledge of SQL correction.

\paragraph{Results on Low-resource Settings.}

\begin{figure}
    \centering
    \includegraphics[width=0.35\textwidth]{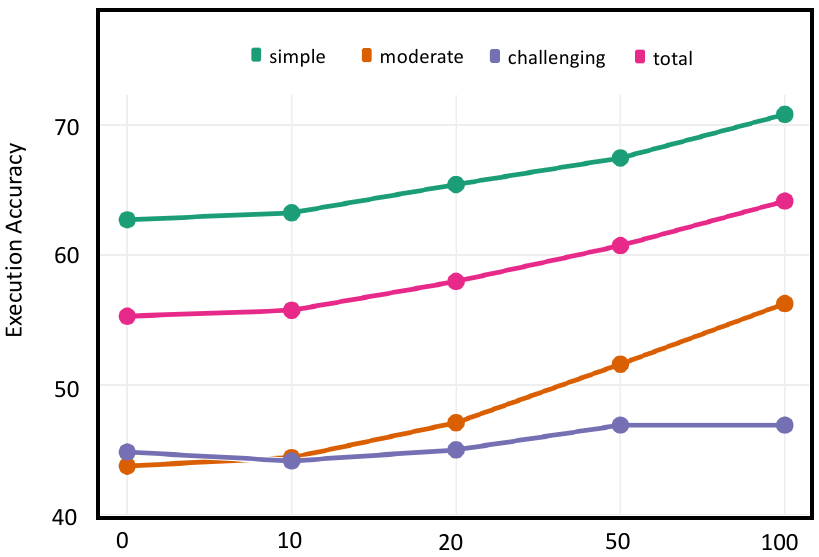}
    \caption{The effect of training data scale on SHARE.}
    \label{fig:data_scale}
\end{figure}

To take a closer look at the impact of the amount and quality of training data on self-correction assistance in LLM, we conduct a low-resource training analysis. Specifically, we sampled three subsets of the training data—10\%, 20\%, and 50\%—and repeated each experiment three times to minimize variance. The averaged results are illustrated in Figure \ref{fig:data_scale}.

Our findings reveal a strong positive correlation between the amount of training data and model performance, thereby confirming the overall high quality of the training set. In particular, GPT-4o + SHARE-8B with only 50\% training data outperforms the state-of-the-art MAGIC baseline, achieving 60.71\% versus MAGIC's 59.53\% in EX in the BIRD data set.

However, this positive trend does not fully hold for more challenging instances. From 0\% to 20\% of the training data, improvements remain unclear and at 10\% there is even a slight decline in performance. Only when the dataset exceeds 20\% of the full training set we do observe a clear performance increase. This suggests that performance improvements for harder instances often remain erratic or limited until the dataset size exceeds a certain point, particularly for tasks involving nuanced or rare examples, as demonstrated by \cite{scale_laws}.

\begin{table}[t]  
    \centering
    \resizebox{0.9\hsize}{!}{
    \begin{tabular}{lcccc}  
    \toprule
    \textbf{\textsc{Method}}& \textbf{\textsc{sim.}} & \textbf{\textsc{mod.}} & \textbf{\textsc{chall.}} & \textbf{\textsc{total}} \\ 
    \midrule
    SHARE-8B &70.81&56.25&46.90& \textbf{64.14}\\
    \midrule
    \textit{(a) w/o Schema Aug} &67.14& 50.22& 46.00& $\textbf{60.02}_\textbf{ ($\downarrow$ 4.08)}$\\
    \textit{(b) w/o Logic Opt} &64.86 & 46.77 & 39.31 & $\textbf{56.98}_\textbf{ ($\downarrow$ 7.16)}$ \\
    \midrule
    \textit{(c) w/o Hierarchy} & 68.34 & 49.89 & 45.02 & $\textbf{60.55}_\textbf{ ($\downarrow$ 3.59)}$ \\
    \textit{(d) w/o Error Pert} &68.08 & 52.79 & 46.20 & $\textbf{61.38}_\textbf{ ($\downarrow$ 2.76)}$ \\
    \bottomrule
    \end{tabular}}
    \caption{Ablation study of SHARE. \textbf{w/o Schema Aug} and \textbf{w/o Logic Opt} denote removing the SAM and LOM, respectively.\textbf{ w/o Hierarchy} denotes training specialized models sequentially instead of employing the hierarchical evolution strategy. \textbf{ w/o Error Pert} denotes the removal of action-based error perturbation.}
    \label{tab:ablation}
\end{table}

\subsection{Ablation Study}
\label{ablation}

Table~\ref{tab:ablation} presents the results of our ablation study. The removal of either the schema or logic refinement module (Table~\ref{tab:ablation} (a–b)) results in substantial performance drops, underscoring the importance of the two-stage refinement architecture for disentangling schema linking from logical reasoning~\citep{sl_role}. Replacing the hierarchical evolution strategy with conventional sequential training (Table~\ref{tab:ablation} (c)) leads to a 3.59\% decline, indicating the advantage of the hierarchical approach in preventing error accumulation and bias transfer across stages~\citep{overfitting}. Furthermore, the exclusion of action-based error perturbation (Table~\ref{tab:ablation} (d)) leads to a 2.76\% reduction, demonstrating the effectiveness of this lightweight augmentation technique. More detailed analysis is provided in Appendix~\ref{app:ablation}.

\subsection{Computational Cost Analysis of SHARE}
\label{sec:cost}
\begin{table*}[t]
    \centering
    \resizebox{0.9 \textwidth}{!}{
    \begin{tabular}{lcccccc}  
    \toprule
    \textbf{\textsc{Method}}& \textbf{\textsc{LLM InToks}} & \textbf{\textsc{LLM OutToks} $\downarrow$} & \textbf{\textsc{SLM InToks } $\downarrow$} & \textbf{\textsc{SLM OutToks} $\downarrow$} & \textbf{\textsc{Cost / 1K} $\downarrow$} & \textbf{\textsc{EX} $\uparrow$}\\ 
    \midrule

     \rowcolor{black!10!}\multicolumn{7}{c}{\textbf{\textit{Inference Stage}}} \\
     MAC-Refiner & \underline{7126.74} & \underline{236.58} & - & - & \underline{\$20.18} & 58.74\\
    Multiple-Prompt &21128.65&1004.55&-&-& \$62.86 &58.80\\
    MAGIC & 8245.16 & 1737.98 & - & - & \$37.99  & \underline{59.53}\\
    GPT-4o + SHARE-8B & \textbf{716.30} & \textbf{68.32} & \textbf{1731.23} & \textbf{132.16} & \textbf{\$2.57}  & \textbf{64.14}\\
    
    \midrule

   \rowcolor{black!10!}\multicolumn{7}{c}{\textbf{\textit{Training Stage}}} \\
    MAGIC & 4838.63 & 2085.22 & - & - & \$32.94  & 59.53\\
    GPT-4o + SHARE-8B & \textbf{1623.28} & \textbf{66.87} & \textbf{2308.75} & \textbf{83.04} & \textbf{\$4.85}  & \textbf{64.14} \\
    
    \bottomrule
    \end{tabular}}
    \caption{Token usage and computational cost of various self-correction methods on BIRD. In(Out)Toks refers to the average input (output) token length per instance. Cost / 1K refers to the average cost per 1000 instances.}
    \label{tab:cost}
\end{table*}

Table \ref{tab:cost} presents the computational cost analysis of SHARE. To the best of our knowledge, our work is the \textbf{first} effort in the text-to-SQL domain to implement correction through collaboration between small and large LMs. This paradigm effectively reduces the inference overhead of SHARE and incurs only \textbf{one-tenth} the cost of the most economical baseline. Notably, SHARE remains highly cost-efficient during the training data construction stage. Compared to the In-Context-Learning-based (ICL-based) training method adopted in MAGIC, which is the strongest self-correction method, our self-evolution strategy significantly reduces the reliance on LLMs during the construction of training data, resulting in substantial cost savings in the overall process. Other relevant details of the cost computation are shown in Appendix \ref{app: cost}.

\subsection{Generalization for Different SQL Dialects}
\label{sec:dialect}
Mainstream text-to-SQL benchmarks predominantly use SQLite as their target SQL dialect, primarily for its accessibility and ease of data collection. However, due to the heightened privacy requirements in data management, MySQL and PostgreSQL, characterized by licensing restrictions and proprietary attributes, are more commonly adopted dialects in real-world implementation scenarios. As demonstrated in Figure \ref{fig:dialect}, SHARE also exhibits effective performance on MySQL and PostgreSQL dialects even without additional training. This can be attributed to SHARE's focus on learning low-level reasoning path corrections, enabling it to generalize across various high-level SQL dialects.

\begin{figure}
    \centering
    \includegraphics[width=0.4\textwidth]{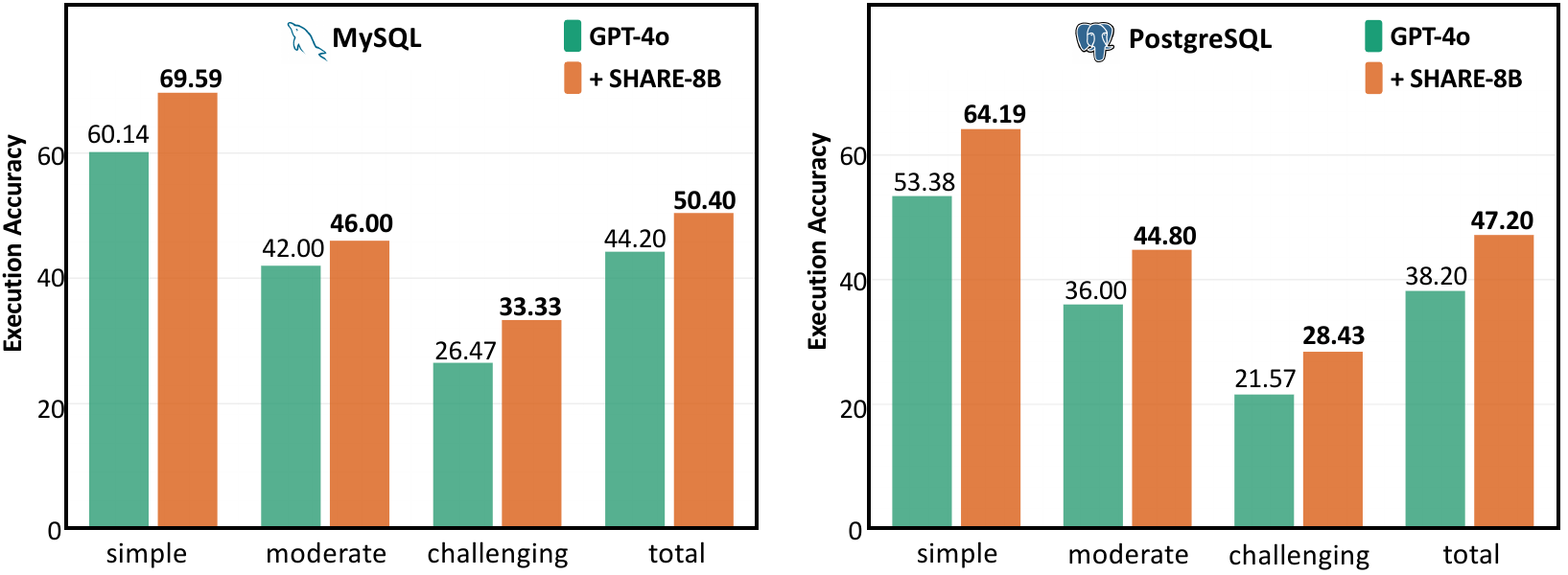}
    \caption{The performance of SHARE on BIRD across various SQL dialects, specifically MySQL (left) and PostgreSQL (right).}
    \label{fig:dialect}
\end{figure}

\subsection{Open-source Teacher Models}
\label{sec:teacher}
\begin{table}[t]  
    \centering
    \resizebox{0.9\hsize}{!}{
    \begin{tabular}{lcccc}  
    \toprule
    \textbf{\textsc{Model}}& \textbf{\textsc{sim.  }} & \textbf{\textsc{mod.  }} & \textbf{\textsc{chall.}} & \textbf{\textsc{total}} \\ 
    \hline
    GPT-4o &63.35&44.18&45.52&55.87\\
    +SHARE-gpt &70.81&56.25&46.90&64.14\\
    +SHARE-llama &71.57&57.54&48.97&65.19\\
    \midrule
    Claude-3.5-S &57.08&39.87&31.03&49.41\\
    +SHARE-gpt &68.86&57.11&50.34&63.56\\
    +SHARE-llama &67.68&52.59&44.83&60.95\\
    \midrule
    Llama-3.1-70B &60.54&44.61&41.38&53.91\\
    +SHARE-gpt &68.86&53.88&43.45&61.93\\
    +SHARE-llama &69.08&56.90&46.90&63.30\\
    \midrule
    Llama-3.1-8B &41.62&27.59&20.00&35.33\\
    +SHARE-gpt &47.68&35.13&28.97&42.11\\
    +SHARE-llama &53.08&35.56&30.34&45.63\\
    \bottomrule
    \end{tabular}}
    \caption{Performance comparison across different teacher models on BIRD.}
    \label{tab:open-source}
  
\end{table}

During the construction of SHARE, as introduced in Section \ref{sec:bam}, GPT-4o acts as the teacher model for automated data synthesis within the Base Action Model (BAM). To strengthen the flexibility and generalization of our workflow, we further investigate the use of open-source teacher models. Specifically, we replace GPT-4o with Llama-3.1-70B to generate training data for the BAM and retrain SHARE-8B accordingly. To clarify the distinction, we refer to the original version of the SHARE-8B model trained with GPT-4o as \textbf{SHARE-gpt}, and the SHARE-8B model trained with Llama-3.1-70B as the teacher model as \textbf{SHARE-llama}. Table \ref{tab:open-source} shows the performance of various generator models assisted by SHARE-gpt and SHARE-llama on the BIRD dev set. It demonstrates that SHARE continues to deliver strong performance even when using an open-source model as the teacher model, and it still works well with a variety of generator models. This suggests that our approach is not restricted by the error patterns of any one model in text-to-SQL, but rather leverages broader knowledge of SQL correction.

\subsection{Quantitative Analysis of SQL Error Corrections}
\begin{figure}
    \centering
    \includegraphics[width=0.3\textwidth]{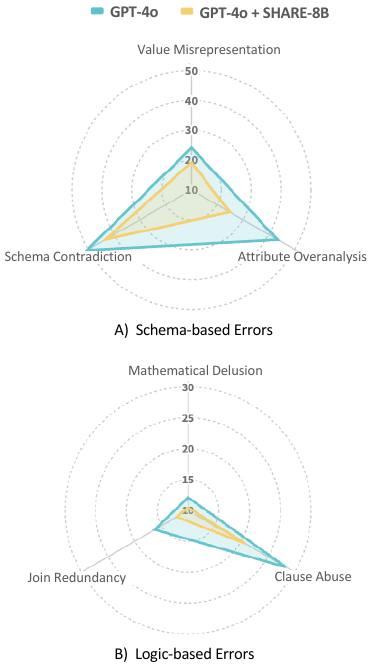}
    \caption{The correction performance across fine-grained error categories on BIRD.}
    \label{fig:errors}
\end{figure}

To quantify the effectiveness of SHARE in error correction, we analyze the SQL queries generated by the GPT-4o baseline. Following TA-SQL \citep{ta-sql}, we categorize the observed errors into two primary types: \textbf{schema-based errors} and \textbf{logic-based errors}. Each category is further subdivided into three specific subtypes. For detailed definitions and examples of these error types, we refer the reader to TA-SQL.

Table \ref{fig:errors} presents the correction performance across fine-grained error categories on the BIRD dev set. The results indicate that SHARE is effective in mitigating both schema-based and logic-based errors, showing substantial reductions in \texttt{Attribute Overanalysis} ($\downarrow$ 18.61\%), \texttt{Schema Contradiction} ($\downarrow$ 7.24\%), and \texttt{Clause Abuse} ($\downarrow$ 7.54\%). However, SHARE demonstrates limited effectiveness in correcting instances of \texttt{Mathematical Delusion} ($\downarrow$ 1.63\%), which may be attributed to the restricted mathematical reasoning capability of the underlying generator model \citep{gym}. To further elucidate SHARE’s correction behavior, we conduct qualitative case studies and present a representative example in Appendix \ref{app:case}

\section{Related Work}

\paragraph{LLMs for Text-to-SQL.}
In recent years, large language models (LLMs) \citep{chatgpt, claude2, gemini} have attracted considerable attention due to their robust reasoning and domain generalization capabilities. The application of LLMs has improved the performance of text-to-SQL to another level of intelligence. Early research leverages in-context learning capabilities of LLMs to develop text-to-SQL systems through meticulously crafted prompt engineering methodologies \citep{evaluate_text_to_sql, din_sql, dail-sql,ta-sql}. With the emergence of language agents \citep{mind2web, middleware} as a promising paradigm, recent works \citep{mac_sql, chess, chase} leverage multi-agent architectures to construct more reliable and comprehensive frameworks for text-to-SQL conversion, yielding substantial improvements in empirical performance. 

\paragraph{Self-correction in Text-to-SQL.}
Self-correction \citep{auto_correct_survey}, where LLMs evaluate and refine their initial output, has emerged as a crucial technology to enhance the reliability and accuracy of automated code generation tasks \citep{cg1, tapilot}. Self-correction has been widely applied in text-to-SQL tasks. Some studies \citep{multiple_prompt, dail-sql} leverage carefully designed prompts to guide LLMs in utilizing their intrinsic reasoning capabilities for more effective self-correction. Additionally, using feedback to effectively guide the self-correction process of LLMs is another promising approach. The sources of feedback are diverse, ranging from human annotations \citep{din_sql}, external execution environments \citep{self-debugging, mac_sql} to iterative exploration conducted by the LLM itself \citep{magic, sql-craft}.

\section{Conclusion}
In this research, we propose SHARE, an SLM-based hierarchical action correction assistant designed to enable more precise error localization and effective self-correction for LLMs. We further propose a novel hierarchical evolution strategy for data-efficient training. Experimental results show the effectiveness and robustness of our method, even in low-source settings, unlocking the potential of SLMs in self-correction for text-to-SQL tasks. 

\newpage
\section{Limitation}
In this paper, we demonstrate the effectiveness of SHARE by presenting the performance of SHARE-assisted self-correction in the one-turn setting, where the generator model receives feedback generated by SHARE and performs a single-revision iteration. We leave investigating the performance of our framework in multi-turn interactive self-correction scenarios, where the correction process undergoes multiple refinement cycles, as our future work. Furthermore, our current work exclusively focuses on the text-to-SQL domain. Expanding SHARE to broader code generation tasks represents another key direction for future research. 

\section{Acknowledgement}
We thank all constructive comments from anonymous reviewers. Reynold Cheng, Ge Qu, Jinyang Li, and Nan Huo are
supported by the Hong Kong Jockey Club Charities Trust
(Project 260920140), the University of Hong Kong (Project
2409100399), the HKU Outstanding Research Student Supervisor Award 2022-23, and the HKU Faculty Exchange
Award 2024 (Faculty of Engineering). Bowen Qin was supported by National Science and Technology Major Project (Project 2022ZD0116306). Chenhao Ma was partially supported by NSFC under Grant 62302421, Basic and Applied Basic Research Fund in Guangdong Province under Grant 2023A1515011280, 2025A1515010439, Ant Group through
CCF-Ant Research Fund, Shenzhen Research Institute of Big Data under grant SIF20240004, and the Guangdong Provincial Key Laboratory of Big Data Computing, The
Chinese University of Hong Kong, Shenzhen. Ge Qu and Jinyang Li were supported by HKU Presidential PhD Scholar Programme. Ge Qu was also funded by Hong Kong PhD Fellowship Scheme.

\section{Ethical Statement}
All datasets employed in this work are publicly accessible. We will also release our models and source code after the review process, ensuring the transparency and reproducibility of our findings. Furthermore, the output generated by our investigations is structured as SQL queries—a programming language format—rather than natural language text, which could potentially involve harmful or biased content. Our team meticulously examines each output to confirm the absence of politically sensitive or biased material. Finally, it is noteworthy that we utilize parameter-efficient LoRA fine-tuning to train our models, which demonstrates superior environmental sustainability compared to full-parameter fine-tuning.

\newpage
\bibliography{anthology,custom}

\begin{thebibliography}{57}
\expandafter\ifx\csname natexlab\endcsname\relax\def\natexlab#1{#1}\fi

\bibitem[{Anthropic(2024)}]{claude2}
Anthropic. 2024.
\newblock \href {https://www.anthropic.com/news/claude-3-haiku} {Claude 3 haiku: our fastest model yet.}

\bibitem[{Askari et~al.(2024)Askari, Poelitz, and Tang}]{magic}
Arian Askari, Christian Poelitz, and Xinye Tang. 2024.
\newblock \href {http://arxiv.org/abs/2406.12692} {Magic: Generating self-correction guideline for in-context text-to-sql}.

\bibitem[{Awan et~al.(2023)Awan, Din, Almogren, and Rodrigues}]{data_privacy}
Kamran~Ahmad Awan, Ikram~Ud Din, Ahmad Almogren, and Joel J. P.~C. Rodrigues. 2023.
\newblock Privacy-preserving big data security for iot with federated learning and cryptography.
\newblock \emph{{IEEE} Access}.

\bibitem[{Berglund et~al.(2024)Berglund, Tong, Kaufmann, Balesni, Stickland, Korbak, and Evans}]{reverse}
Lukas Berglund, Meg Tong, Maximilian Kaufmann, Mikita Balesni, Asa~Cooper Stickland, Tomasz Korbak, and Owain Evans. 2024.
\newblock The reversal curse: Llms trained on "a is b" fail to learn "b is a".
\newblock In \emph{The Twelfth International Conference on Learning Representations, {ICLR} 2024, Vienna, Austria, May 7-11, 2024}.

\bibitem[{Chen et~al.(2023)Chen, Zhang, Nguyen, Zan, Lin, Lou, and Chen}]{codet}
Bei Chen, Fengji Zhang, Anh Nguyen, Daoguang Zan, Zeqi Lin, Jian{-}Guang Lou, and Weizhu Chen. 2023.
\newblock Codet: Code generation with generated tests.
\newblock In \emph{The Eleventh International Conference on Learning Representations, {ICLR} 2023, Kigali, Rwanda, May 1-5, 2023}.

\bibitem[{Chen et~al.(2024{\natexlab{a}})Chen, Lin, Sch{\"{a}}rli, and Zhou}]{self-debugging}
Xinyun Chen, Maxwell Lin, Nathanael Sch{\"{a}}rli, and Denny Zhou. 2024{\natexlab{a}}.
\newblock Teaching large language models to self-debug.
\newblock In \emph{The Twelfth International Conference on Learning Representations, {ICLR} 2024, Vienna, Austria, May 7-11, 2024}.

\bibitem[{Chen et~al.(2024{\natexlab{b}})Chen, Lin, Sch{\"{a}}rli, and Zhou}]{self-debug}
Xinyun Chen, Maxwell Lin, Nathanael Sch{\"{a}}rli, and Denny Zhou. 2024{\natexlab{b}}.
\newblock Teaching large language models to self-debug.
\newblock In \emph{The Twelfth International Conference on Learning Representations, {ICLR} 2024, Vienna, Austria, May 7-11, 2024}.

\bibitem[{Deng et~al.(2024)Deng, Gu, Zheng, Chen, Stevens, Wang, Sun, and Su}]{mind2web}
Xiang Deng, Yu~Gu, Boyuan Zheng, Shijie Chen, Sam Stevens, Boshi Wang, Huan Sun, and Yu~Su. 2024.
\newblock Mind2web: Towards a generalist agent for the web.
\newblock \emph{Advances in Neural Information Processing Systems}.

\bibitem[{Dou et~al.(2023)Dou, Gao, Pan, Wang, Che, Lou, and Zhan}]{uni_sar}
Longxu Dou, Yan Gao, Mingyang Pan, Dingzirui Wang, Wanxiang Che, Jian{-}Guang Lou, and Dechen Zhan. 2023.
\newblock Unisar: a unified structure-aware autoregressive language model for text-to-sql semantic parsing.
\newblock \emph{Int. J. Mach. Learn. Cybern.}

\bibitem[{Dziri et~al.(2021)Dziri, Madotto, Za{\"{\i}}ane, and Bose}]{dialog_ha_1}
Nouha Dziri, Andrea Madotto, Osmar Za{\"{\i}}ane, and Avishek~Joey Bose. 2021.
\newblock Neural path hunter: Reducing hallucination in dialogue systems via path grounding.
\newblock In \emph{Proceedings of the 2021 Conference on Empirical Methods in Natural Language Processing, {EMNLP} 2021, Virtual Event / Punta Cana, Dominican Republic, 7-11 November, 2021}.

\bibitem[{Gan et~al.(2021)Gan, Chen, and Purver}]{dk}
Yujian Gan, Xinyun Chen, and Matthew Purver. 2021.
\newblock Exploring underexplored limitations of cross-domain text-to-sql generalization.
\newblock In \emph{Proceedings of the 2021 Conference on Empirical Methods in Natural Language Processing, {EMNLP} 2021, Virtual Event / Punta Cana, Dominican Republic, 7-11 November, 2021}.

\bibitem[{Gao et~al.(2024)Gao, Wang, Li, Sun, Qian, Ding, and Zhou}]{dail-sql}
Dawei Gao, Haibin Wang, Yaliang Li, Xiuyu Sun, Yichen Qian, Bolin Ding, and Jingren Zhou. 2024.
\newblock Text-to-sql empowered by large language models: {A} benchmark evaluation.
\newblock \emph{Proc. {VLDB} Endow.}

\bibitem[{Gou et~al.(2024)Gou, Shao, Gong, Shen, Yang, Duan, and Chen}]{critic}
Zhibin Gou, Zhihong Shao, Yeyun Gong, Yelong Shen, Yujiu Yang, Nan Duan, and Weizhu Chen. 2024.
\newblock {CRITIC:} large language models can self-correct with tool-interactive critiquing.
\newblock In \emph{The Twelfth International Conference on Learning Representations, {ICLR} 2024, Vienna, Austria, May 7-11, 2024}.

\bibitem[{Gu(2023)}]{cg1}
Qiuhan Gu. 2023.
\newblock Llm-based code generation method for golang compiler testing.
\newblock In \emph{Proceedings of the 31st {ACM} Joint European Software Engineering Conference and Symposium on the Foundations of Software Engineering, {ESEC/FSE} 2023, San Francisco, CA, USA, December 3-9, 2023}.

\bibitem[{Gu et~al.(2024)Gu, Shu, Yu, Liu, Dong, Tang, Srinivasa, Latapie, and Su}]{middleware}
Yu~Gu, Yiheng Shu, Hao Yu, Xiao Liu, Yuxiao Dong, Jie Tang, Jayanth Srinivasa, Hugo Latapie, and Yu~Su. 2024.
\newblock Middleware for llms: Tools are instrumental for language agents in complex environments.
\newblock \emph{arXiv preprint arXiv:2402.14672}.

\bibitem[{Hu et~al.(2022)Hu, Shen, Wallis, Allen{-}Zhu, Li, Wang, Wang, and Chen}]{pure_lora}
Edward~J. Hu, Yelong Shen, Phillip Wallis, Zeyuan Allen{-}Zhu, Yuanzhi Li, Shean Wang, Lu~Wang, and Weizhu Chen. 2022.
\newblock Lora: Low-rank adaptation of large language models.
\newblock In \emph{The Tenth International Conference on Learning Representations, {ICLR} 2022, Virtual Event, April 25-29, 2022}.

\bibitem[{Huang et~al.(2024)Huang, Chen, Mishra, Zheng, Yu, Song, and Zhou}]{cannot_self_correct}
Jie Huang, Xinyun Chen, Swaroop Mishra, Huaixiu~Steven Zheng, Adams~Wei Yu, Xinying Song, and Denny Zhou. 2024.
\newblock Large language models cannot self-correct reasoning yet.
\newblock In \emph{The Twelfth International Conference on Learning Representations, {ICLR} 2024, Vienna, Austria, May 7-11, 2024}.

\bibitem[{Ji et~al.(2023)Ji, Lee, Frieske, Yu, Su, Xu, Ishii, Bang, Madotto, and Fung}]{ha_nlg}
Ziwei Ji, Nayeon Lee, Rita Frieske, Tiezheng Yu, Dan Su, Yan Xu, Etsuko Ishii, Yejin Bang, Andrea Madotto, and Pascale Fung. 2023.
\newblock Survey of hallucination in natural language generation.
\newblock \emph{{ACM} Comput. Surv.}

\bibitem[{Kaplan et~al.(2020)Kaplan, McCandlish, Henighan, Brown, Chess, Child, Gray, Radford, Wu, and Amodei}]{scale_laws}
Jared Kaplan, Sam McCandlish, Tom Henighan, Tom~B. Brown, Benjamin Chess, Rewon Child, Scott Gray, Alec Radford, Jeffrey Wu, and Dario Amodei. 2020.
\newblock Scaling laws for neural language models.

\bibitem[{Lee et~al.(2024)Lee, Park, Kim, and Park}]{multiple_prompt}
Dongjun Lee, Choongwon Park, Jaehyuk Kim, and Heesoo Park. 2024.
\newblock \href {http://arxiv.org/abs/2405.07467} {Mcs-sql: Leveraging multiple prompts and multiple-choice selection for text-to-sql generation}.

\bibitem[{Lei et~al.(2020)Lei, Wang, Ma, Gan, Lu, Kan, and Chua}]{sl_role}
Wenqiang Lei, Weixin Wang, Zhixin Ma, Tian Gan, Wei Lu, Min-Yen Kan, and Tat-Seng Chua. 2020.
\newblock Re-examining the role of schema linking in text-to-{SQL}.
\newblock In \emph{Proceedings of the 2020 Conference on Empirical Methods in Natural Language Processing (EMNLP)}.

\bibitem[{Li et~al.(2023)Li, Hui, Cheng, Qin, Ma, Huo, Huang, Du, Si, and Li}]{graphix}
Jinyang Li, Binyuan Hui, Reynold Cheng, Bowen Qin, Chenhao Ma, Nan Huo, Fei Huang, Wenyu Du, Luo Si, and Yongbin Li. 2023.
\newblock Graphix-t5: Mixing pre-trained transformers with graph-aware layers for text-to-sql parsing.
\newblock In \emph{Thirty-Seventh {AAAI} Conference on Artificial Intelligence, {AAAI} 2023, Thirty-Fifth Conference on Innovative Applications of Artificial Intelligence, {IAAI} 2023, Thirteenth Symposium on Educational Advances in Artificial Intelligence, {EAAI} 2023, Washington, DC, USA, February 7-14, 2023}.

\bibitem[{Li et~al.(2024{\natexlab{a}})Li, Hui, Qu, Yang, Li, Li, Wang, Qin, Geng, Huo et~al.}]{bird}
Jinyang Li, Binyuan Hui, Ge~Qu, Jiaxi Yang, Binhua Li, Bowen Li, Bailin Wang, Bowen Qin, Ruiying Geng, Nan Huo, et~al. 2024{\natexlab{a}}.
\newblock Can llm already serve as a database interface? a big bench for large-scale database grounded text-to-sqls.
\newblock \emph{Advances in Neural Information Processing Systems}.

\bibitem[{Li et~al.(2024{\natexlab{b}})Li, Huo, Gao, Shi, Zhao, Qu, Wu, Ma, Lou, and Cheng}]{tapilot}
Jinyang Li, Nan Huo, Yan Gao, Jiayi Shi, Yingxiu Zhao, Ge~Qu, Yurong Wu, Chenhao Ma, Jian-Guang Lou, and Reynold Cheng. 2024{\natexlab{b}}.
\newblock Tapilot-crossing: Benchmarking and evolving llms towards interactive data analysis agents.

\bibitem[{Li and Xie(2024)}]{right_sql}
Zhenwen Li and Tao Xie. 2024.
\newblock Using llm to select the right sql query from candidates.

\bibitem[{Liu et~al.(2021)Liu, Yang, and Wang}]{overfitting}
Huihui Liu, Yiding Yang, and Xinchao Wang. 2021.
\newblock Overcoming catastrophic forgetting in graph neural networks.
\newblock In \emph{Thirty-Fifth {AAAI} Conference on Artificial Intelligence, {AAAI} 2021, Thirty-Third Conference on Innovative Applications of Artificial Intelligence, {IAAI} 2021, The Eleventh Symposium on Educational Advances in Artificial Intelligence, {EAAI} 2021, Virtual Event, February 2-9, 2021}.

\bibitem[{Liu and Tan(2024)}]{epi-sql}
Xiping Liu and Zhao Tan. 2024.
\newblock \href {http://arxiv.org/abs/2404.14453} {Epi-sql: Enhancing text-to-sql translation with error-prevention instructions}.

\bibitem[{Mirzadeh et~al.(2025)Mirzadeh, Alizadeh, Shahrokhi, Tuzel, Bengio, and Farajtabar}]{gym}
Seyed{-}Iman Mirzadeh, Keivan Alizadeh, Hooman Shahrokhi, Oncel Tuzel, Samy Bengio, and Mehrdad Farajtabar. 2025.
\newblock Gsm-symbolic: Understanding the limitations of mathematical reasoning in large language models.
\newblock In \emph{The Thirteenth International Conference on Learning Representations, {ICLR} 2025, Singapore, April 24-28, 2025}.

\bibitem[{Ouyang et~al.(2022{\natexlab{a}})Ouyang, Wu, Jiang, Almeida, Wainwright, Mishkin, Zhang, Agarwal, Slama, Ray, Schulman, Hilton, Kelton, Miller, Simens, Askell, Welinder, Christiano, Leike, and Lowe}]{follow_feedback}
Long Ouyang, Jeffrey Wu, Xu~Jiang, Diogo Almeida, Carroll~L. Wainwright, Pamela Mishkin, Chong Zhang, Sandhini Agarwal, Katarina Slama, Alex Ray, John Schulman, Jacob Hilton, Fraser Kelton, Luke Miller, Maddie Simens, Amanda Askell, Peter Welinder, Paul~F. Christiano, Jan Leike, and Ryan Lowe. 2022{\natexlab{a}}.
\newblock Training language models to follow instructions with human feedback.

\bibitem[{Ouyang et~al.(2022{\natexlab{b}})Ouyang, Wu, Jiang, Almeida, Wainwright, Mishkin, Zhang, Agarwal, Slama, Ray, Schulman, Hilton, Kelton, Miller, Simens, Askell, Welinder, Christiano, Leike, and Lowe}]{chatgpt}
Long Ouyang, Jeffrey Wu, Xu~Jiang, Diogo Almeida, Carroll~L. Wainwright, Pamela Mishkin, Chong Zhang, Sandhini Agarwal, Katarina Slama, Alex Ray, John Schulman, Jacob Hilton, Fraser Kelton, Luke Miller, Maddie Simens, Amanda Askell, Peter Welinder, Paul~F. Christiano, Jan Leike, and Ryan Lowe. 2022{\natexlab{b}}.
\newblock Training language models to follow instructions with human feedback.
\newblock In \emph{Advances in Neural Information Processing Systems 35: Annual Conference on Neural Information Processing Systems 2022, NeurIPS 2022, New Orleans, LA, USA, November 28 - December 9, 2022}.

\bibitem[{Pan et~al.(2024{\natexlab{a}})Pan, Saxon, Xu, Nathani, Wang, and Wang}]{automatic_correction_survey}
Liangming Pan, Michael Saxon, Wenda Xu, Deepak Nathani, Xinyi Wang, and William~Yang Wang. 2024{\natexlab{a}}.
\newblock Automatically correcting large language models: \emph{Surveying the Landscape of Diverse Automated Correction Strategies}.
\newblock \emph{Trans. Assoc. Comput. Linguistics}.

\bibitem[{Pan et~al.(2024{\natexlab{b}})Pan, Saxon, Xu, Nathani, Wang, and Wang}]{auto_correct_survey}
Liangming Pan, Michael Saxon, Wenda Xu, Deepak Nathani, Xinyi Wang, and William~Yang Wang. 2024{\natexlab{b}}.
\newblock Automatically correcting large language models: Surveying the landscape of diverse automated correction strategies.
\newblock \emph{Transactions of the Association for Computational Linguistics}.

\bibitem[{Peng and Zhang(2024)}]{enhanced_kd}
Tianyu Peng and Jiajun Zhang. 2024.
\newblock \href {http://arxiv.org/abs/2409.12545} {Enhancing knowledge distillation of large language models through efficient multi-modal distribution alignment}.

\bibitem[{Pourreza et~al.(2024)Pourreza, Li, Sun, Chung, Talaei, Kakkar, Gan, Saberi, Ozcan, and Arik}]{chase}
Mohammadreza Pourreza, Hailong Li, Ruoxi Sun, Yeounoh Chung, Shayan Talaei, Gaurav~Tarlok Kakkar, Yu~Gan, Amin Saberi, Fatma Ozcan, and Sercan~O. Arik. 2024.
\newblock Chase-sql: Multi-path reasoning and preference optimized candidate selection in text-to-sql.

\bibitem[{Pourreza and Rafiei(2024)}]{din_sql}
Mohammadreza Pourreza and Davood Rafiei. 2024.
\newblock Din-sql: Decomposed in-context learning of text-to-sql with self-correction.
\newblock \emph{Advances in Neural Information Processing Systems}.

\bibitem[{Qin et~al.(2022)Qin, Wang, Hui, Li, Wei, Li, Huang, Si, Yang, and Li}]{sun}
Bowen Qin, Lihan Wang, Binyuan Hui, Bowen Li, Xiangpeng Wei, Binhua Li, Fei Huang, Luo Si, Min Yang, and Yongbin Li. 2022.
\newblock {SUN:} exploring intrinsic uncertainties in text-to-sql parsers.
\newblock In \emph{Proceedings of the 29th International Conference on Computational Linguistics, {COLING} 2022, Gyeongju, Republic of Korea, October 12-17, 2022}.

\bibitem[{Qiu et~al.(2020)Qiu, Sun, Xu, Shao, Dai, and Huang}]{plm}
XiPeng Qiu, TianXiang Sun, YiGe Xu, YunFan Shao, Ning Dai, and XuanJing Huang. 2020.
\newblock Pre-trained models for natural language processing: A survey.
\newblock \emph{Science China Technological Sciences}.

\bibitem[{Qu et~al.(2024)Qu, Li, Li, Qin, Huo, Ma, and Cheng}]{ta-sql}
Ge~Qu, Jinyang Li, Bowen Li, Bowen Qin, Nan Huo, Chenhao Ma, and Reynold Cheng. 2024.
\newblock Before generation, align it! {A} novel and effective strategy for mitigating hallucinations in text-to-sql generation.
\newblock Association for Computational Linguistics.

\bibitem[{Rafailov et~al.(2023)Rafailov, Sharma, Mitchell, Manning, Ermon, and Finn}]{dpo}
Rafael Rafailov, Archit Sharma, Eric Mitchell, Christopher~D. Manning, Stefano Ermon, and Chelsea Finn. 2023.
\newblock Direct preference optimization: Your language model is secretly a reward model.
\newblock In \emph{Advances in Neural Information Processing Systems 36: Annual Conference on Neural Information Processing Systems 2023, NeurIPS 2023, New Orleans, LA, USA, December 10 - 16, 2023}.

\bibitem[{Rajkumar et~al.(2022)Rajkumar, Li, and Bahdanau}]{evaluate_text_to_sql}
Nitarshan Rajkumar, Raymond Li, and Dzmitry Bahdanau. 2022.
\newblock Evaluating the text-to-sql capabilities of large language models.

\bibitem[{Sivasubramaniam et~al.(2024)Sivasubramaniam, Osei-Akoto, Zhang, Stockinger, and Fuerst}]{sm3}
Sithursan Sivasubramaniam, Cedric Osei-Akoto, Yi~Zhang, Kurt Stockinger, and Jonathan Fuerst. 2024.
\newblock Sm3-text-to-query: Synthetic multi-model medical text-to-query benchmark.

\bibitem[{Talaei et~al.(2024)Talaei, Pourreza, Chang, Mirhoseini, and Saberi}]{chess}
Shayan Talaei, Mohammadreza Pourreza, Yu-Chen Chang, Azalia Mirhoseini, and Amin Saberi. 2024.
\newblock Chess: Contextual harnessing for efficient sql synthesis.

\bibitem[{Team et~al.(2024)Team, Anil, Borgeaud, Alayrac, Yu, Soricut, Schalkwyk, and M}]{gemini}
Gemini Team, Rohan Anil, Sebastian Borgeaud, Jean-Baptiste Alayrac, Jiahui Yu, Radu Soricut, Johan Schalkwyk, and Andrew M. 2024.
\newblock Gemini: A family of highly capable multimodal models.

\bibitem[{Wang et~al.(2024)Wang, Ren, Yang, Liang, Bai, Chai, Yan, Zhang, Yin, Sun, and Li}]{mac_sql}
Bing Wang, Changyu Ren, Jian Yang, Xinnian Liang, Jiaqi Bai, Linzheng Chai, Zhao Yan, Qian-Wen Zhang, Di~Yin, Xing Sun, and Zhoujun Li. 2024.
\newblock \href {http://arxiv.org/abs/2312.11242} {Mac-sql: A multi-agent collaborative framework for text-to-sql}.

\bibitem[{Wang et~al.(2023)Wang, Wei, Schuurmans, Le, Chi, Narang, Chowdhery, and Zhou}]{self_consistency}
Xuezhi Wang, Jason Wei, Dale Schuurmans, Quoc~V. Le, Ed~H. Chi, Sharan Narang, Aakanksha Chowdhery, and Denny Zhou. 2023.
\newblock Self-consistency improves chain of thought reasoning in language models.
\newblock In \emph{The Eleventh International Conference on Learning Representations, {ICLR} 2023, Kigali, Rwanda, May 1-5, 2023}.

\bibitem[{Xia et~al.(2024)Xia, Jiang, Deng, Wang, Zhao, Mihalcea, and Zhang}]{sql-craft}
Hanchen Xia, Feng Jiang, Naihao Deng, Cunxiang Wang, Guojiang Zhao, Rada Mihalcea, and Yue Zhang. 2024.
\newblock $r^3$: "this is my sql, are you with me?" a consensus-based multi-agent system for text-to-sql tasks.

\bibitem[{Xie et~al.(2024)Xie, Jin, Xie, Matrixmxlin, Chen, Yu, Lei, Zhuo, Hu, and Li}]{decomposition_t2s}
Yuanzhen Xie, Xinzhou Jin, Tao Xie, Matrixmxlin Matrixmxlin, Liang Chen, Chenyun Yu, Cheng Lei, Chengxiang Zhuo, Bo~Hu, and Zang Li. 2024.
\newblock Decomposition for enhancing attention: Improving llm-based text-to-sql through workflow paradigm.
\newblock In \emph{Findings of the Association for Computational Linguistics, {ACL} 2024, Bangkok, Thailand and virtual meeting, August 11-16, 2024}.

\bibitem[{Xu et~al.(2024{\natexlab{a}})Xu, Fu, Gao, Ye, Liu, Mei, Wang, Yu, and Wu}]{dpo_better}
Shusheng Xu, Wei Fu, Jiaxuan Gao, Wenjie Ye, Weilin Liu, Zhiyu Mei, Guangju Wang, Chao Yu, and Yi~Wu. 2024{\natexlab{a}}.
\newblock Is dpo superior to ppo for llm alignment? a comprehensive study.
\newblock In \emph{Proceedings of the 41st International Conference on Machine Learning}.

\bibitem[{Xu et~al.(2024{\natexlab{b}})Xu, Li, Tao, Shen, Cheng, Li, Xu, Tao, and Zhou}]{kd_survey}
Xiaohan Xu, Ming Li, Chongyang Tao, Tao Shen, Reynold Cheng, Jinyang Li, Can Xu, Dacheng Tao, and Tianyi Zhou. 2024{\natexlab{b}}.
\newblock A survey on knowledge distillation of large language models.

\bibitem[{Yin and Neubig(2017)}]{logical_synthesis}
Pengcheng Yin and Graham Neubig. 2017.
\newblock A syntactic neural model for general-purpose code generation.
\newblock In \emph{Proceedings of the 55th Annual Meeting of the Association for Computational Linguistics, {ACL} 2017, Vancouver, Canada, July 30 - August 4, Volume 1: Long Papers}.

\bibitem[{Yu et~al.(2018)Yu, Zhang, Yang, Yasunaga, Wang, Li, Ma, Li, Yao, Roman, Zhang, and Radev}]{spider}
Tao Yu, Rui Zhang, Kai Yang, Michihiro Yasunaga, Dongxu Wang, Zifan Li, James Ma, Irene Li, Qingning Yao, Shanelle Roman, Zilin Zhang, and Dragomir~R. Radev. 2018.
\newblock Spider: {A} large-scale human-labeled dataset for complex and cross-domain semantic parsing and text-to-sql task.
\newblock In \emph{Proceedings of the 2018 Conference on Empirical Methods in Natural Language Processing, Brussels, Belgium, October 31 - November 4, 2018}.

\bibitem[{Zhang et~al.(2024)Zhang, Lan, Zhu, Liu, Hoang, Kokane, Yao, Tan, Prabhakar, Chen, Liu, Feng, Awalgaonkar, Murthy, Hu, Chen, Xu, Niebles, Heinecke, Wang, Savarese, and Xiong}]{action_model}
Jianguo Zhang, Tian Lan, Ming Zhu, Zuxin Liu, Thai Hoang, Shirley Kokane, Weiran Yao, Juntao Tan, Akshara Prabhakar, Haolin Chen, Zhiwei Liu, Yihao Feng, Tulika Awalgaonkar, Rithesh Murthy, Eric Hu, Zeyuan Chen, Ran Xu, Juan~Carlos Niebles, Shelby Heinecke, Huan Wang, Silvio Savarese, and Caiming Xiong. 2024.
\newblock \href {http://arxiv.org/abs/2409.03215} {xlam: A family of large action models to empower ai agent systems}.

\bibitem[{Zhang et~al.(2025)Zhang, Wang, Dou, Zhu, and Che}]{table_reasoning}
Xuanliang Zhang, Dingzirui Wang, Longxu Dou, Qingfu Zhu, and Wanxiang Che. 2025.
\newblock A survey of table reasoning with large language models.
\newblock \emph{Frontiers Comput. Sci.}

\bibitem[{Zhao et~al.(2024)Zhao, Zhou, Li, Tang, Wang, Hou, Min, Zhang, Zhang, Dong, Du, Yang, Chen, Chen, Jiang, Ren, Li, Tang, Liu, Liu, Nie, and Wen}]{llm_survey}
Wayne~Xin Zhao, Kun Zhou, Junyi Li, Tianyi Tang, Xiaolei Wang, Yupeng Hou, Yingqian Min, Beichen Zhang, Junjie Zhang, Zican Dong, Yifan Du, Chen Yang, Yushuo Chen, Zhipeng Chen, Jinhao Jiang, Ruiyang Ren, Yifan Li, Xinyu Tang, Zikang Liu, Peiyu Liu, Jian-Yun Nie, and Ji-Rong Wen. 2024.
\newblock A survey of large language models.

\bibitem[{Zhao et~al.(2025)Zhao, Luo, Shi, Chen, Wang, Che, Liu, and Sun}]{chart}
Xuanle Zhao, Xianzhen Luo, Qi~Shi, Chi Chen, Shuo Wang, Wanxiang Che, Zhiyuan Liu, and Maosong Sun. 2025.
\newblock Chartcoder: Advancing multimodal large language model for chart-to-code generation.

\bibitem[{Zheng et~al.(2024)Zheng, Zhang, Zhang, Ye, and Luo}]{llamafactory}
Yaowei Zheng, Richong Zhang, Junhao Zhang, Yanhan Ye, and Zheyan Luo. 2024.
\newblock {L}lama{F}actory: Unified efficient fine-tuning of 100+ language models.
\newblock In \emph{Proceedings of the 62nd Annual Meeting of the Association for Computational Linguistics (Volume 3: System Demonstrations)}. Association for Computational Linguistics.

\bibitem[{Zhong et~al.(2023)Zhong, Snell, Klein, and Eisner}]{case_study_t2s}
Ruiqi Zhong, Charlie Snell, Dan Klein, and Jason Eisner. 2023.
\newblock Non-programmers can label programs indirectly via active examples: {A} case study with text-to-sql.
\newblock In \emph{Proceedings of the 2023 Conference on Empirical Methods in Natural Language Processing, {EMNLP} 2023, Singapore, December 6-10, 2023}, pages 5126--5152.

\end{thebibliography}
\bibliographystyle{acl_natbib}

\clearpage
\newpage

\appendix
\section{SHARE Recipe}

\subsection{Training Data Distribution}
\label{app:training_data_dist}
\begin{table}
    \centering
    \resizebox{0.9\hsize}{!}{
    \begin{tabular}{lcc}  
    \toprule
    \textbf{\textsc{Model}}& \textbf{\textsc{\#Input}} & \textbf{\textsc{\#Output}} \\ 
    \midrule
    Base Action Model &196.23&25.71\\
    Schema Augmentation Model &621.46&30.24\\
    Logic Optimation Model  &683.69&29.80 \\
    \bottomrule
    \end{tabular}}
    \caption{The average length of input and output tokens in the training corpus for each model.}
    \label{tab:tokens}
\end{table}

In this work, we utilize the training set of two mainstream cross-domain text-to-SQL benchmarks as the seed data to construct our own training data. 1) \textbf{\textsc{BIRD}}: The training set of BIRD contains 9,428 pairs of text-to-SQL  data and 69 big databases across 37 professional domains. 2) \textbf{\textsc{SPIDER}}: SPIDER is a more standard text-to-SQL benchmark that contains 8659 training examples across more than 20 domains. Table \ref{tab:tokens} displays the detailed average length of the input and output tokens for each model in our own training data. Our approach achieved superior performance with notably few training tokens, validating the data-efficient nature of SHARE. 

\begin{table}
    \centering
    \resizebox{0.9\hsize}{!}{
    \begin{tabular}{lcc}  
    \toprule
    \textbf{\textsc{Shorthand Name}}& \textbf{\textsc{Official Model Alias
}} \\ 
    \midrule
    Llama-3.1-70B &\texttt{Llama-3.1-70B-Instruct}\\
    Qwen-Coder-32B  & \texttt{Qwen2.5-Coder-32B-Instruct} \\
    Llama-3.1-8B & \texttt{Llama-3.1-8B-Instruct}\\
    DS-Coder-6.7B  & \texttt{deepseek-coder-6.7b-instruct} \\
    \bottomrule
    \end{tabular}}
    \caption{Shorthand model names used throughout the paper and their corresponding official model aliases.}
    \label{tab:full_name}
\end{table}

\begin{table}
    \centering
    \resizebox{0.9\hsize}{!}{
    \begin{tabular}{lcccc}  
    \toprule
    \textbf{\textsc{Approach}}& \textbf{\textsc{Simple}} & \textbf{\textsc{Moderate}}
    & \textbf{\textsc{Challenging}} & \textbf{\textsc{Total}}\\ 
    \midrule
    DPO & 65.08 &48.06 &42.76&57.82\\
    LoRA &\textbf{70.81} & \textbf{56.25} & \textbf{46.90} & \textbf{64.14} \\
    \bottomrule
    \end{tabular}}
    \caption{The performance of SHARE trained via different fine-tuning approaches on BIRD dev.}
    \label{tab:dpo}
\end{table}

\subsection{Action Space of SHARE}
\label{app:actions}
As discussed in Section \ref{sec:bam}, we design actions in target trajectories as pandas-like APIs and employ GPT-4o to convert each ground truth SQL to a verified action trajectory by few-shot prompting for the training data construction. All logically meaningful and validated actions generated by GPT-4o in this process are collected as the action space of SHARE. We further categorize these actions into four types, including clause, dataframe, aggregation, and operator types. Specific actions are presented in Table \ref{tab:action_1} and \ref{tab:action_2}.

\begin{table*}[t]
\centering
\footnotesize
\resizebox{1.0\linewidth}{!}{
\begin{tabular}{L{2cm}p{1.5cm}p{3cm}p{7.5cm}}
\toprule
\textbf{Category} & \textbf{Action} & \textbf{Expression} & \textbf{Explanation}\\
\midrule

\multirow{0}{*}{Clause} 
& \texttt{SELECT} & select(elements) &
Select data from the database.
\newline
\textbf{Parameters:} \newline
\textcolor{teal}{\textbf{elements}} - the selected elements. Valid elements include qualified column names in the format of \texttt{table.column}, aggregate functions, or any valid SQL-syntax selectable entity. Multiple elements are separated by commas.\\

\cmidrule{2-4}

& \texttt{WHERE} & where(element, filter) &
Filter rows by conditions.
\newline
\textbf{Parameters:} \newline
\textcolor{teal}{\textbf{elements}} - The qualified column, or expression to be filtered.
\newline
\textcolor{teal}{\textbf{filter}} - The condition that determines which values from the element are included.\\

\cmidrule{2-4}

& \texttt{GROUP BY} & groupby(elements) &
Groups rows that have the same values into summary rows.
\newline
\textbf{Parameters:} \newline
\textcolor{teal}{\textbf{elements}} - Qualified columns, or expressions used to group the data. Multiple elements are separated by commas.\\

\cmidrule{2-4}
& \texttt{HAVEING} & having(element, filter) &
Filter groups of data with aggregate functions. 
\newline
\textbf{Parameters:} \newline
\textcolor{teal}{\textbf{elements}} - Qualified columns, or expressions used to filter data. Multiple elements are separated by commas. 
\newline
\textcolor{teal}{\textbf{filter}} - The condition that determines which values from the element are included.\\

\cmidrule{2-4}
& \texttt{ORDER BY} & orderby(by, order) &
Sort the result set based on a qualified column or expression.
\newline
\textbf{Parameters:} \newline
\textcolor{teal}{\textbf{by}} - The qualified column or expression used to sort the data. 
\newline
\textcolor{teal}{\textbf{order}} - The sorted order. It should be \texttt{DESC} or \texttt{ASC}. \\

\cmidrule{2-4}
& \texttt{LIMIT} & limit(num) &
Restrict the number of rows returned. 
\newline
\textbf{Parameters:} \newline
\textcolor{teal}{\textbf{num}} - The num is a flexible input that specifies the type of limitation to apply. For instance, \texttt{limit(1)} denotes limiting the number of rows to 1.  \texttt{limit(2,9)} denotes specifying a range of columns to return, which is columns 2 through 9. \\

\cmidrule{2-4}
& \texttt{DISTINCT} & distinct(element) &
Remove duplicate rows from the result set. 
\newline
\textbf{Parameters:} \newline
\textcolor{teal}{\textbf{element}} - The qualified column to be processed. \\
                    
\midrule

\multirow{0}{*}{Dataframe} &
\texttt{UNION} & df1.union(df2) & 
Union the result set of two dataframes. \\
\cmidrule{2-4}
& \texttt{INTERSECT} & df1.intersect(df2) & 
Intersect the result set of two dataframes. \\     
\cmidrule{2-4}
& \texttt{EXCEPT} & df1.except(df2) & 
Subtract the result of df2 from the result of df1. \\ 

\midrule
\multirow{0}{*}{Aggregation} &
\texttt{SUM} & sum(element) & 
Sum all non-null values of the element or expression. 
\newline
\textbf{Parameters:} \newline
\textcolor{teal}{\textbf{element}} - The target qualified column or expression. \\

\cmidrule{2-4}
&\texttt{AVG} & average(element) & 
Calculate the average of all non-null values for the element or expression. 
\newline
\textbf{Parameters:} \newline
\textcolor{teal}{\textbf{element}} - The target qualified column or expression. \\

\cmidrule{2-4}
&\texttt{COUNT} & count(element) & 
Returns the total number of the element or expression. 
\newline
\textbf{Parameters:} \newline
\textcolor{teal}{\textbf{element}} - The target qualified column or expression. \\

\cmidrule{2-4}
&\texttt{MIN} & min(element) & 
Minimum value in the element or expression except for nULL. 
\newline
\textbf{Parameters:} \newline
\textcolor{teal}{\textbf{element}} - The target qualified column or expression. \\

\cmidrule{2-4}
&\texttt{MAX} & max(element) & 
Maximum value in the element or expression except for nULL. 
\newline
\textbf{Parameters:} \newline
\textcolor{teal}{\textbf{element}} - The target qualified column or expression. \\

\bottomrule
\end{tabular}}
\caption{Actions categorized as clause, dataframe, and aggregation types employed in the construction of action trajectories within SHARE. Actions categorized as operator types are shown in Table \ref{tab:action_2}.}
\label{tab:action_1}
\end{table*}

\begin{table*}[t]
\centering
\footnotesize
\resizebox{1.0\linewidth}{!}{
\begin{tabular}{L{2cm}p{1.5cm}p{3cm}p{7.5cm}}
\toprule
\textbf{Category} & \textbf{Action} & \textbf{Expression} & \textbf{Explanation}\\
\midrule
\multirow{0}{*}{Operator} 
&\texttt{CAST} & cast(element, type) & 
Convert the element or expression to the target data type.
\newline
\textbf{Parameters:} \newline
\textcolor{teal}{\textbf{element}} - The qualified column or expression to convert. \newline
\textcolor{teal}{\textbf{type}} - The target data type.\\

\cmidrule{2-4}
&\texttt{CALCU- \newline LATION} & +, -, *, / & 
Implement corresponding calculations for two expressions.  \\

\cmidrule{2-4}
&\texttt{SUBSTR} & substr(element, \newline piv, len) & 
Extract a substring, starting from a specified position and optionally for a specified length 
\newline
\textbf{Parameters:} \newline
\textcolor{teal}{\textbf{element}} - The qualified column to extract. \newline
\textcolor{teal}{\textbf{piv}} - The start point. \newline
\textcolor{teal}{\textbf{len}} - The extracted length. \\

\bottomrule
\end{tabular}}
\caption{Actions categorized as operator actions employed in the construction of action trajectories within SHARE. Actions categorized as other types are shown in Table \ref{tab:action_1}.}
\label{tab:action_2}
\end{table*}

\subsection{Erroneous Trajectory Collection}
\label{app:ps_code}
As introduced in Section \ref{sec:lom}, the erroneous action trajectories in the training data for LOM come from two resources: 1) initial erroneous SQLs, and 2) verified action trajectories that are perturbed by our action-based perturbation strategy. To better clarify the data augmentation through error perturbations, we present the pseudocode of this process in Pseudocode \ref{code}. 


\begin{algorithm}[h!]
    \caption{Data Augmentation}
    \begin{algorithmic}[1]
        \State \textbf{Inputs:} \\
        $verified\_corrs$
        \Comment{A collection of verified correct trajectories  {$verified\_corr\_1$,  $verified\_corr\_2$, …}} \\
        $SLM$
        \Comment{Small Language Model}\\
        $K$
        \Comment{Number of error perturbations per ground truth}
        \State \textbf{Output:} \\
        $data\_pairs$
        \Comment{A set of (erroneous trajectory, verified correct trajectory) pairs}
        \Statex
        \State \textbf{Initialize:}
        \State Initialize $data\_pairs$ as an empty set.
        \Statex

        \For{$verified\_corr\_i$ \textbf{in} $verified\_corrs$} 
            \For{$p = 1$ to $K$}
                \State $er\_p \gets \text{ErrorPerturb}(verified\_corr\_i, SLM)$
                \Comment{Use the smaller model to inject errors into the verified correct trajectory $verified\_corr\_i$.}
                \State Add the pair $(er\_p, verified\_corr\_i)$ to $data\_pairs$
                \Comment{No additional verification is needed, since $verified\_corr\_i$ is correct by definition.}
            \EndFor
        \EndFor
        
        \State \Return $data\_pairs$
    
    \end{algorithmic}
    \label{code}
\end{algorithm}

\section{Experiment Setup}

\subsection{Baseline Methods}
\label{app:baselines}

Table \ref{tab:mainresult} presents all the baseline methods we use for comparison. In this work, we consider Self-Correction \citep{cannot_self_correct}, Self-Consistency \citep{self_consistency}, Multiple-Prompt \citep{multiple_prompt} as feedback-independent self-correction baselines. These methods leverage the intrinsic capabilities of the LLM through prompt engineering to enable its self-correction. For feedback-dependent self-correction baselines, we implement Self-Debugging \citep{self-debugging}, DIN-Correction \citep{din_sql}, MAC-Refiner \citep{mac_sql}, and MAGIC \citep{magic}, where the Large Language Model (LLM) refines its initial output under the guidance of feedback. Feedback utilized in these approaches is derived from three primary sources: human annotations, external execution environments, and through LLMs iteratively exploring contextual environments.

The details of these methods are as follows: 1) \textbf{Self-Correction} is a naive method in which the LLM reconsiders and refines its outputs through vanilla Chain-of-Thought (CoT) prompting. 2) \textbf{Self-Consistency} \citep{self_consistency} is a method that refines the initial output by exploring a broader search space and selecting the most consistent one. We generate five SQL queries using the baseline SQL generation prompt implemented in BIRD \citep{bird} and consider an instance to be correctly solved if at least one of the generated SQL queries produces the correct result. 3) \textbf{Multiple-Prompt} generates diverse queries by systematically reordering candidate tables within the prompt. Following the implementation of \citep{multiple_prompt}, we generate up to five combinations of prompts for each instance and employ the same evaluation mechanism as self-consistency to determine the results. 4) \textbf{Self-Debugging} \citep{self-debug} generates the feedback by investing the execution results and explaining the generated SQL in natural language. The feedback further serves as guidance to instruct the LLM to self-correct.  5) \textbf{DIN-Correction} utilizes the human-annotated guideline from DIN-SQL \citep{din_sql} for self-correction. 6) \textbf{MAC-Refiner}, which is a sub-agent of MAC-SQL \citep{mac_sql}, implements the self-correction process based on multi-dimensional error feedback information by analyzing the execution results, including syntactic correctness, execution feasibility, and retrieval of non-empty results. 7) \textbf{Magic} \citep{magic} collaborates on the failure experiences and automatically distills correction guidelines, employing a crafted LLM-based multi-agent framework for self-correction.

\subsection{Hyper-Parameters}
\label{app:hyper_para}
We set the low-rank dimensions as 8, the learning rate as $5e^{-5}$, and the batch size as 8. We train 5 epochs for the Base Action Model (BAM) and train 3 epochs for the Schema Augmentation Model (SAM) and the Logic Optimization Model (LOM). During inference, we set the temperature as 0.1, the top p as 0.95, and the maximum sample length as 1024. We report the experimental results as the average of five repeated trials.

\subsection{Model Reference Mapping}
\label{app:full_name}
We list in Table \ref{tab:full_name} the shorthand model names used throughout the paper alongside their corresponding official model aliases.

\section{Further Anaylsis}

\subsection{Ablation Study}
\label{app:ablation}
Table \ref{tab:ablation} presents the results of our ablation study, aimed at isolating and evaluating contributions of each component of SHARE. As shown in Table \ref{tab:ablation}(a)-(b), the substantial performance degradation resulting from the removal of either refinement action model underscores the importance of our \textbf{two-stage refinement} design. This two-stage approach, which separately handles schema linking and logical reasoning, proves to be essential for effective text-to-SQL correction \citep{sl_role}.

To further examine the benefits of the \textbf{hierarchical evolution strategy} in SHARE, we compare it against a sequential training pipeline reminiscent of classical continual learning. When trained sequentially, performance declines by 3.59\% (Table \ref{tab:ablation}(c)), suggesting that later models may be disproportionately influenced by biases or errors carried over from earlier training stages \citep{overfitting}.  In contrast, SHARE employs hierarchical evolution during training and strategically integrates knowledge at inference time, mitigating these limitations and achieving superior results.

Finally, Table \ref{tab:ablation}(d) highlights the effectiveness of \textbf{action-based error perturbation} as a data augmentation strategy. Although simple, it yields a 2.76\% improvement in performance, reinforcing its value as a straightforward yet potent enhancement to SHARE’s overall text-to-SQL reasoning capability.

\subsection{Computational Cost of SHARE}
\label{app: cost}

As illustrated in Section \ref{sec:cost}, we compare the average token usage per instance and average computational cost per 1000 instances during the inference and training stages for GPT-4o assisted by SHARE, versus other strong LLM-based text-to-SQL correction approaches on the BIRD development set. We present the usage of input and output tokens separately since they have different prices. We take the same price for the calculation as in the previous work \citep{sm3} \footnote{Pricing of GPT-4o API: \href{https://openai.com/api/pricing/}{https://openai.com/api/pricing/}. Price of Llama-3.1 8B usage: \href{https://groq.com/pricing/}{https://groq.com/pricing/}}. 

Notably, SHARE remains highly cost-efficient during the training data construction stage. As introduced in Section \ref{sec:bam}, SHARE just uses GPT-4o to generate and verify the training data for the Base Action Model (BAM). Afterward, BAM itself creates the training data for all other models, enabling a self-evolution process without further reliance on GPT-4o. To more clearly demonstrate SHARE’s cost-efficiency in training, we compare it against the strongest self-correction method, MAGIC. While MAGIC does not undergo a direct fine-tuning step, it adopts an ICL-based training approach, which follows a training-like procedure where the LLM explores the training set, produces a correction guideline for each instance, and memorizes successful corrections as task knowledge for inference. As shown in Table \ref{tab:cost}, the sharp reduction in token usage directly translates to significant computational savings, underscoring SHARE’s superior cost-efficiency in the training stage.

\subsection{Independent Inference of SHARE}
While SHARE shows clear improvements among various generator LLMs, \textbf{a key question} is how well it performs independently, without any external LLM grounding. This question is especially important for privacy-sensitive scenarios, such as text-to-SQL tasks on confidential relational databases.

To address this, we evaluate SHARE’s standalone capabilities, using the same orchestration prompting strategy as for \texttt{Llama-3.1-8B Orchestration}. As shown in Figure~\ref{fig:grounding}, SHARE-8B achieves robust performance on its own, surpassing both \texttt{Llama-3.1-8B} and \texttt{Llama-3.1-8B Orchestration}, and approaching the quality of strong proprietary models like GPT-35-Turbo, Claude-3.5-Sonnet, and GPT-4o-mini. 

This improvement results from our parameter-efficient LoRa fine-tuning approach, which augments the model’s capabilities without altering its original parameters. By optimizing only a small set of new parameters, we enhance the underlying SLM’s text-to-SQL performance without compromising its existing strengths. Moreover, this efficient design supports secure on-device deployment that does not require exposing sensitive database content, relying solely on an 8B-scale SLM and a modest set of additional parameters.

\begin{figure}
    \centering
    \includegraphics[width=0.4\textwidth]{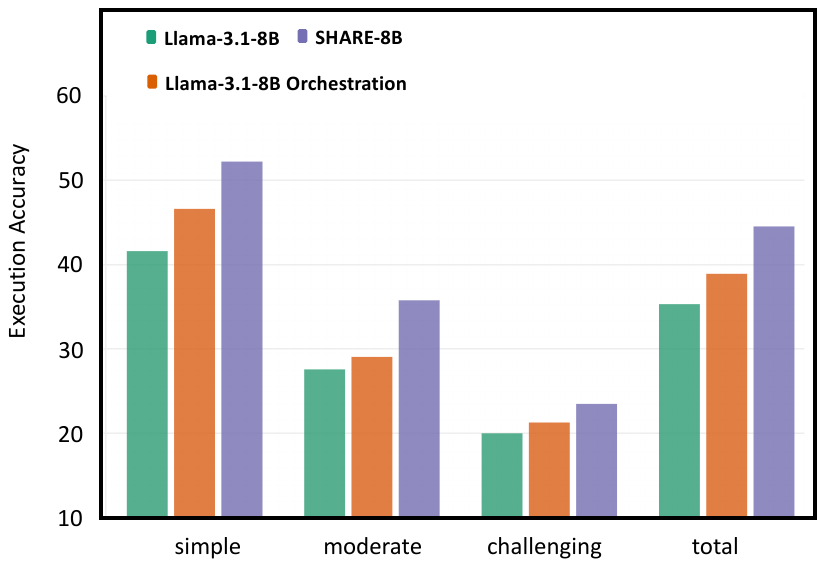}
    \caption{Independent inference performance of SHARE on BIRD.}
    \label{fig:grounding}
\end{figure}

\subsection{Analysis of Overcorrections}
Overcorrection \citep{automatic_correction_survey}, which refers to the modification of initially correct SQL queries into incorrect ones, is a notorious challenge in self-correction. To mitigate this issue, we tailored the training data for the Logic Optimization Model (LOM). Specifically, we consider pairs of (erroneous trajectory, verified trajectory) as positive samples and, at a 4:1 ratio, introduce (verified trajectory, verified trajectory) pairs as negative samples.

Analysis of the final correction results shows that, compared to the current leading self-correction method, MAGIC, SHARE reduced the overcorrection rate from \textbf{15.52\%} to \textbf{11.20\%}. This demonstrates that our design mitigates overfitting in the correction process in a simple but effective way, preventing the SLMs from treating all trajectories as erroneous and unnecessarily fixing them.

\subsection{Fine-tuning Approach Exploration}
Apart from LoRA, we also explored training the SLMs in SHARE using Direct Preference Optimization (DPO) \citep{dpo} during our initial attempt. In the data construction stage, verified action trajectories are considered as the chosen responses. We employ the action-based perturbation strategy introduced in Section \ref{sec:lom} and provide 5-shot examples to guide GPT-4o to generate rejected responses based on chosen responses. 

As shown in Table \ref{tab:dpo}, compared to SHARE-8B trained via LoRA, the performance gains of the DPO-trained version showed a consistent decline across all difficulty levels. We observe that the DPO model outputs many meaningless trajectories, such as invalid pandas output format. We speculate that the suboptimal performance of DPO-based training arises from its limited suitability for low-resource settings. Furthermore, for data science code generation tasks \citep{chart} that require complex reasoning, DPO relies on high-quality data and requires crafted rejected responses by human annotation \citep{dpo_better}. In contrast, LoRA can effectively leverage automatically generated data, making it more efficient under these constraints. Consequently, we adopted LoRA as our primary fine-tuning method, given its lightweight design and effectiveness. 

\subsection{Case Study}
\label{app:case}
\begin{table*}[t]  
\centering
\small
\scalebox{0.8}{
\begin{tabular}{llc}
\toprule
\rowcolor{blue!15!}  & \textbf{\textcolor{black}{A Detailed Case}}& \\ 
 \midrule
 \leavevmode {\bf \color{blue}Question} & \textit{Which county reported the most number of school closures in the 1980s with school ownership code} \\ 
 & \textit{belonging to Youth Authority Facilities (CEA)?} &  \\
 \leavevmode {\bf \color{blue}Evidence} & \textit{Youth Authority Facilities (CEA) refers to SOC = 11; 1980s = years between 1980 and 1989.} \\ 
 \leavevmode {\bf \color{blue}Gold} & SELECT {\color{blue}County} FROM schools WHERE strftime('\%Y', {\color{blue}ClosedDate}) BETWEEN '1980' AND '1989'   & \\
 & AND SOC = 11 GROUP BY County ORDER BY {\color{blue}COUNT(ClosedDate)} DESC LIMIT 1
 &  \\ 

 \hdashline
  \leavevmode {\bf \color{red}Initial SQL} & SELECT County, {\color{red}COUNT(*)} AS YearCount FROM schools WHERE {\color{red}Year} BETWEEN '1980-01-01'&  \\ 
 & AND '1989-12-31' AND SOC = 11 GROUP BY County ORDER BY YearCount DESC LIMIT 1; &  \\
 \hdashline
  \leavevmode {\bf \color{orange}MAGIC} & SELECT County, {\color{red}COUNT(*)} as ClosureCount FROM schools WHERE {\color{teal}ClosedDate} BETWEEN '1980-01-01'& \\ 
 & AND '1989-12-31' AND SOC = 11 GROUP BY County ORDER BY ClosureCount DESC LIMIT 1; &  \\
 \hdashline
 \leavevmode {\bf \color{teal}SHARE} & SELECT {\color{teal}schools.County} FROM schools WHERE schools.SOC = 11 AND {\color{teal}schools.ClosedDate} BETWEEN '1980-01-01'&  \\ 
 & AND '1989-12-31' GROUP BY schools.County ORDER BY COUNT(schools.ClosedDate) DESC LIMIT 1; &  \\

 \midrule
 \rowcolor{blue!15!} &\textbf{\textcolor{black}{Output Log
}} & \\ 
  \midrule
 \leavevmode {\bf \color{red}Initial SQL} & SELECT County, {\color{red}COUNT(*)} AS YearCount FROM schools WHERE {\color{red}Year} BETWEEN '1980-01-01'&  \\ 
 & AND '1989-12-31' AND SOC = 11 GROUP BY County ORDER BY YearCount DESC LIMIT 1; &  \\
 \hdashline
  \leavevmode {\bf \color{teal} BAM:} & \texttt{df1 = df.where(element = {\color{red}schools.Year}, filter = 'between 1980-01-01 and 1989-12-31'))} &  \\
 & \texttt{df2 = df1.where(element = schools.SOC, filter = 11)}
&\\
&\texttt{df3 = df2.groupby(schools.County).{\color{red}count(schools.Year)}}
&\\
&\texttt{df4 = df3.orderby(by = {\color{red}count(schools.Year)}, desc).limit(1)}
&\\
&\texttt{res = df4.select({\color{red}schools.County, count(schools.Year)})}
&\\ 

  \hdashline
  \leavevmode {\bf \color{teal} SAM:} & \texttt{df1 = df.where(element = {\color{teal}schools.ClosedDate}, filter = 'between 1980-01-01 and 1989-12-31'))} &  \\
 & \texttt{df2 = df1.where(element = schools.SOC, filter = 11)}
&\\
&\texttt{df3 = df2.groupby(schools.County).{\color{teal}count(schools.ClosedDate)}}
&\\
&\texttt{df4 = df3.orderby(by = {\color{teal}count(schools.ClosedDate)}, desc).limit(1)}
&\\
&\texttt{res = df4.select({\color{red}schools.County, count(schools.ClosedDate)})}
&\\ 

\hdashline
  \leavevmode {\bf \color{teal} LOM:} & \texttt{df1 = df.where(element = {\color{teal}schools.ClosedDate}, filter = 'between 1980-01-01 and 1989-12-31'))} &  \\
 & \texttt{df2 = df1.where(element = schools.SOC, filter = 11)}
&\\
&\texttt{df3 = df2.groupby(schools.County).{\color{teal}count(schools.ClosedDate)}}
&\\
&\texttt{df4 = df3.orderby(by = {\color{teal}count(schools.ClosedDate)}, desc).limit(1)}
&\\
&\texttt{res = df4.select({\color{teal}schools.County})}
&\\ 

 \hdashline
 \leavevmode {\bf \color{teal}Final Output} & SELECT schools.County FROM schools WHERE schools.SOC = 11 AND schools.ClosedDate BETWEEN '1980-01-01'&  \\ 
 & AND '1989-12-31' GROUP BY schools.County ORDER BY COUNT(schools.ClosedDate) DESC LIMIT 1; &  \\

 \bottomrule
\end{tabular}}
\caption{Case study: an illustrative case from BIRD dev.}
\label{tab:case}
\end{table*}

In order to provide deeper insights into SHARE’s effectiveness, we conducted case studies on its correction outputs. Table \ref{tab:case} presents an illustrative case randomly selected from the BIRD development set. By decomposing the declarative SQL query into an action trajectory and leveraging the multi-SLM orchestration during inference, SHARE achieves more granular corrections. Consequently, it produces a correct query that effectively addresses both schema and logic issues. We additionally offer an output log that contains the output of each model during SHARE inference for reference.

\section{Implement Prompts}

\subsection{Data Construction for BAM}
\label{app:bam_data_prompt}

\begin{figure*}[t]
    \centering
    \includegraphics[width=1.0\textwidth]{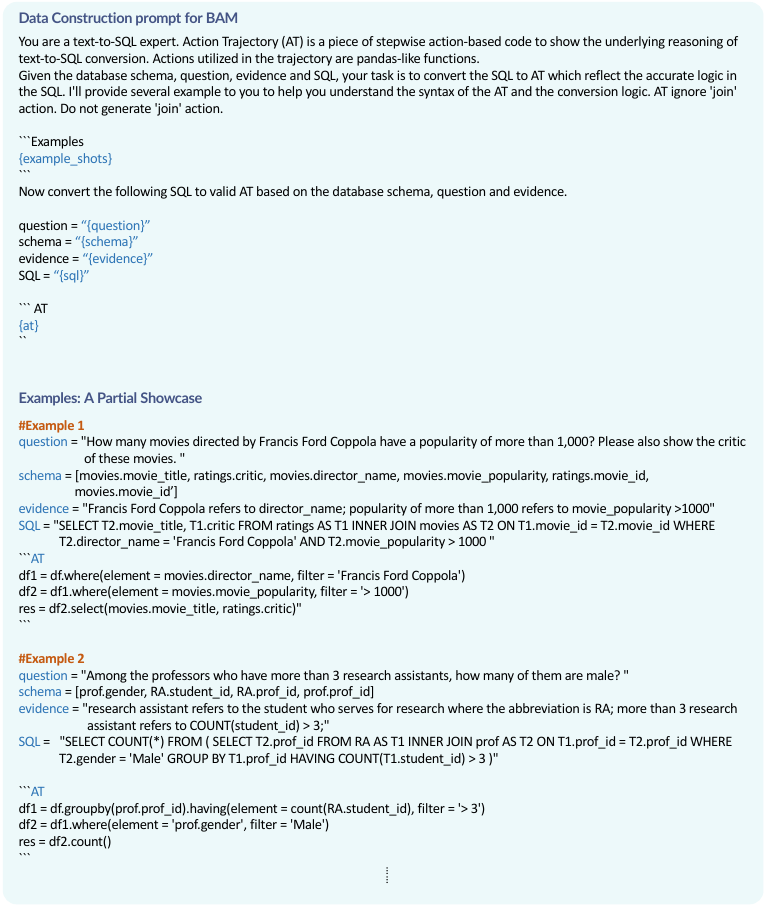}
    \caption{The prompt for the data construction process of BAM.}
    \label{fig:bam_prompt}
\end{figure*}

Figure \ref{fig:bam_prompt} illustrates the prompts employed in our data construction process for BAM. Specifically, we utilize seven examples to guide GPT-4o in generating corresponding action trajectories for SQL queries. The figure presents two representative examples, while the remaining examples can be found in our code.

\subsection{Error Perturbation Prompt}

\begin{figure*}[t]
    \centering
    \includegraphics[width=1.0\textwidth]{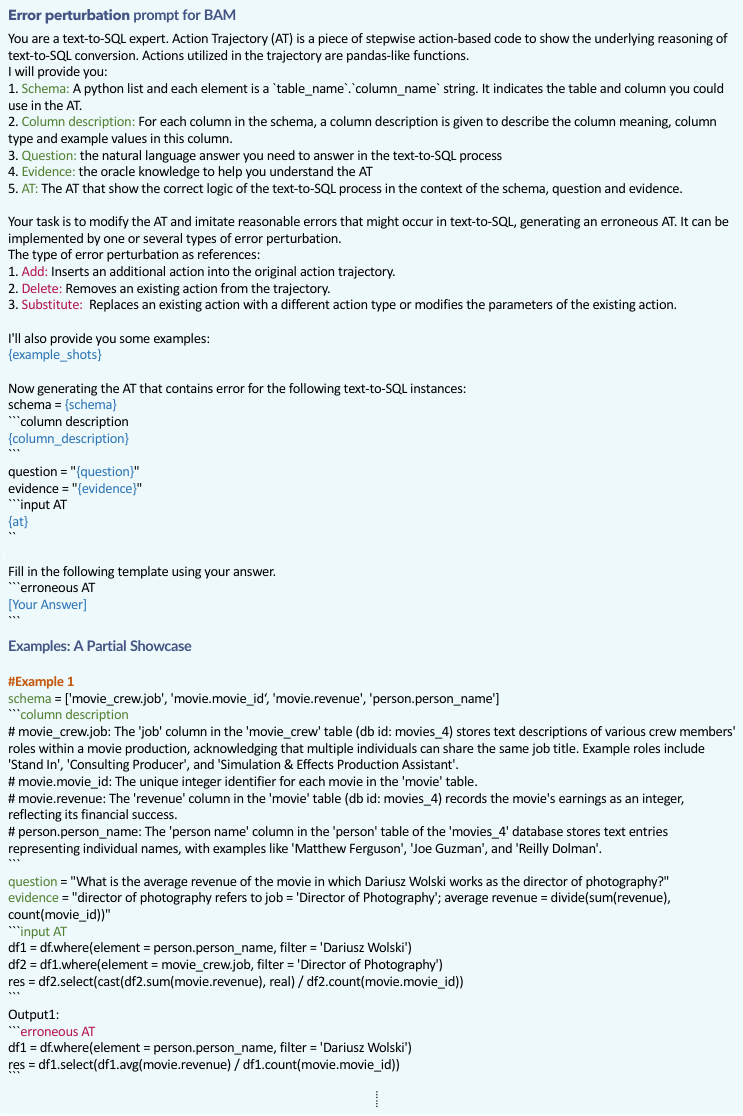}
    \caption{The prompt for the error perturbation strategy implemented by BAM.}
    \label{fig:error_prompt}
\end{figure*}

Figure \ref{fig:error_prompt} illustrates the prompts we use to implement error perturbations during the data construction process of LOM. Specifically, we utilize four examples to guide BAM to generate perturbed action trajectories. The figure presents a representative example, while the remaining examples can be found in our code.

\subsection{Inference Prompts}
\label{app:bam_data_prompt}

\begin{figure*}[t]
    \centering
    \includegraphics[width=1.0\textwidth]{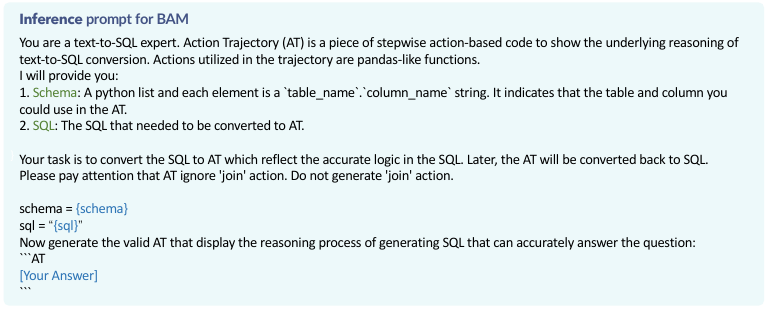}
    \caption{The prompt for BAM to convert SQL to action trajectory.}
    \label{fig:bam_infer}
\end{figure*}

\begin{figure*}[t]
    \centering
    \includegraphics[width=1.0\textwidth]{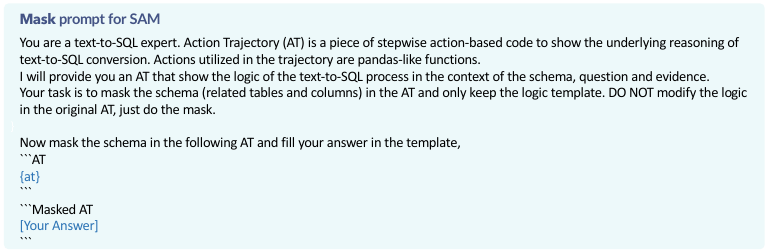}
    \caption{The prompt for SAM to generate the schema-masked variant given the input trajectory.}
    \label{fig:sam_mask}
\end{figure*}

\begin{figure*}[t]
    \centering
    \includegraphics[width=1.0\textwidth]{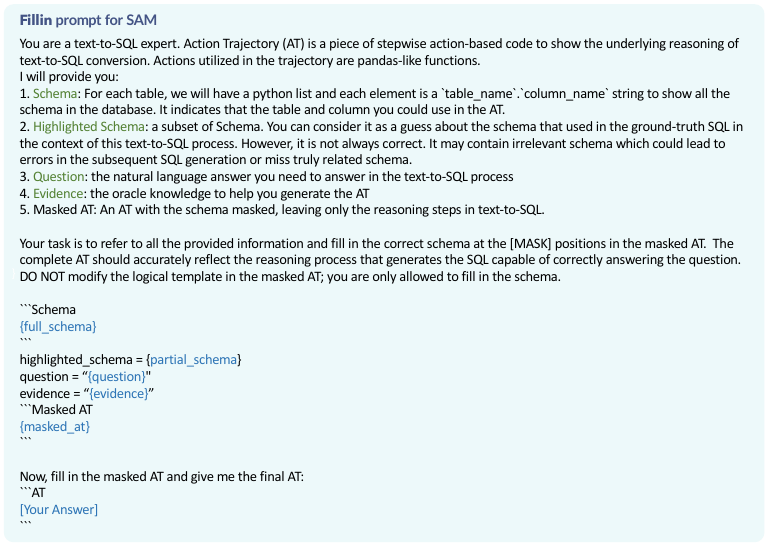}
    \caption{The prompt for SAM to reinsert the correct schema in the schema-based variant.}
    \label{fig:sam_fill}
\end{figure*}

\begin{figure*}[t]
    \centering
    \includegraphics[width=1.0\textwidth]{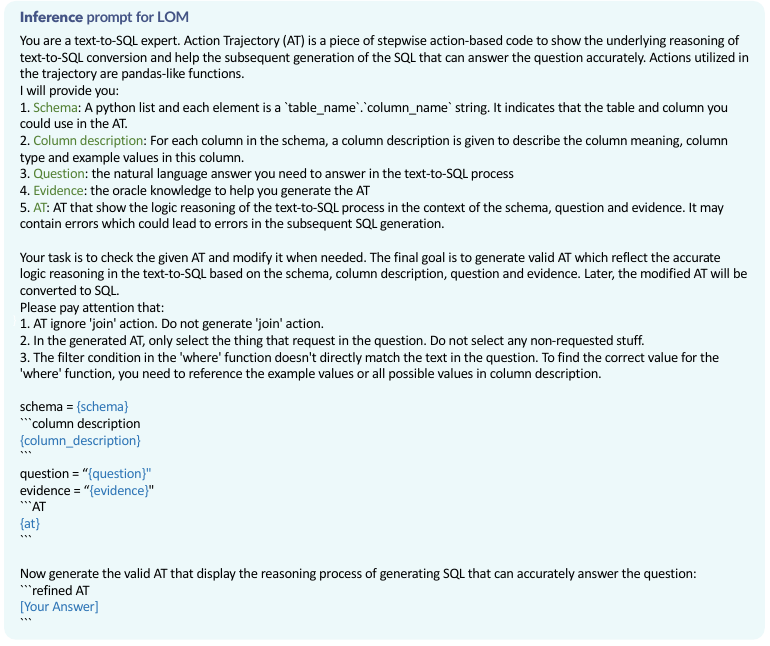}
    \caption{The prompt for LOM to rectify logic-related errors in the input trajectory.}
    \label{fig:lom_infer}
\end{figure*}

\begin{figure*}[t]
    \centering
    \includegraphics[width=1.0\textwidth]{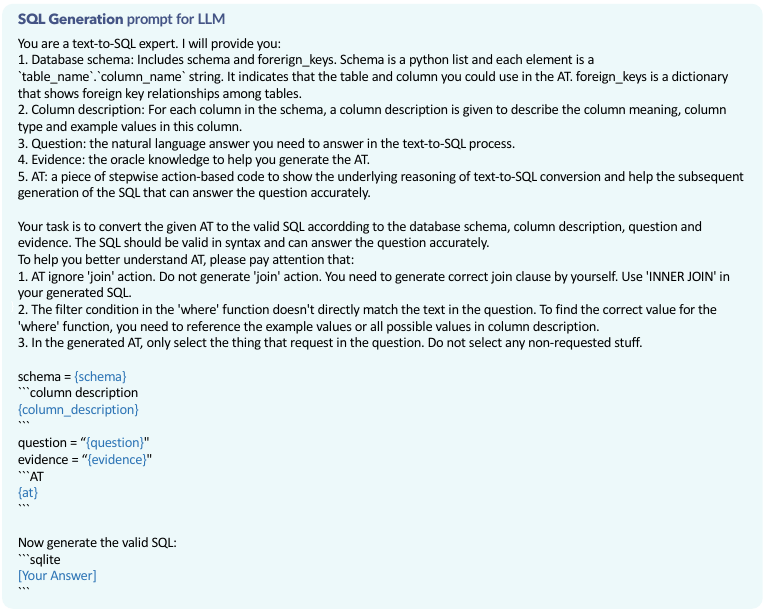}
    \caption{The prompt for LLM to generate refined SQL given the trajectory outputted by SHARE.}
    \label{fig:llm_sql_generation}
\end{figure*}

In this section, we present prompts that we use in the inference process of SHARE. During inference, we prompt BAM to convert the initial SQL using the prompt in Figure \ref{fig:bam_infer}. Subsequently, the generated trajectory is forwarded to SAM. The SAM first masks the schema using the prompt in Figure \ref{fig:sam_mask} and then generates the schema-refined trajectory using the prompt in Figure \ref{fig:sam_fill}. The LOM then takes this trajectory as input and refines it as the final output of SHARE using the prompt in Figure \ref{fig:lom_infer}. Finally, the SHARE-produced trajectory serves as feedback to facilitate the self-correction of LLM, as shown in Figure \ref{fig:llm_sql_generation}. All the prompts are in the zero-shot setting.

\section{Reproducebility}
\label{app:reproduce}
We fine-tune and infer open-source models, including \texttt{Llama-3.1-8b-Instruct} and \texttt{Phi-3-mini-4K-instruct} (3.8B) via \texttt{LlamaFactory}\footnote{https://github.com/hiyouga/LLaMA-Factory}. The \texttt{Llama-3.1-70B-Instruct} model is inferred using \texttt{vllm}\footnote{https://github.com/vllm-project/vllm}. To accelerate its inference process, we also implemented \texttt{deepspeed}\footnote{https://github.com/microsoft/DeepSpeed}. All open-source models are accessed via \texttt{huggingface}\footnote{https://huggingface.co/models}.

We will open-source the source code along with the training data, model checkpoints, and prompts in each stage after the anonymous review phase.

\end{document}